\documentclass[10pt,twocolumn,letterpaper]{article}
\pdfoutput=1

\usepackage{iccv}
\usepackage{times}
\usepackage{epsfig}
\usepackage{graphicx}
\usepackage{amsmath}
\usepackage{amssymb}
\usepackage{math}

\usepackage{multirow}
\usepackage{xspace}
\usepackage{pifont}
\usepackage{xcolor}
\usepackage{textcomp}
\usepackage{dsfont}
\usepackage{booktabs}
\usepackage{tabularx}

\usepackage{enumitem}%
\usepackage{rotating}
\usepackage{bbm}
\usepackage{float}
\usepackage{stmaryrd} %

\def\mypar#1{\vspace{1mm}{\noindent\bf #1.}\hspace{1mm}}
\def\sect#1{Sec.~\ref{sec:#1}}
\def\Eq#1{Eq.~(\ref{eq:#1})}
\def\fig#1{Fig.~\ref{fig:#1}}
\def\tab#1{Tab.~\ref{tab:#1}}
\def\myvspace{\vspace{1mm}} %

\usepackage[pagebackref=true,breaklinks=true,colorlinks,bookmarks=false]{hyperref}

\definecolor{myred}{RGB}{246,37,135}
\definecolor{myyel}{RGB}{228,201,32}
\definecolor{myblue}{RGB}{35,175,170}

\newcommand{\red}[1]{\textbf{\textcolor{myred}{#1}}}
\newcommand{\blue}[1]{\textbf{\textcolor{myblue}{#1}}}
\newcommand{\yel}[1]{\textbf{\textcolor{myyel}{#1}}}

\newcommand{\reds}[1]{\textbf{\textbf{\textcolor{myred}{#1}}}}
\newcommand{\blues}[1]{\textbf{\textbf{\textcolor{myblue}{#1}}}}
\newcommand{\yels}[1]{\textbf{\textbf{\textcolor{myyel}{#1}}}}

\newcommand{\cmark}{\textcolor{green!80!black}{\ding{51}}}
\newcommand{\xmark}{\textcolor{red}{\ding{55}}}

 \iccvfinalcopy %

\def\iccvPaperID{4195} %

\ificcvfinal\fi

\begin{document}

\def \ours {{ZestGuide}\xspace}
\title{Zero-shot spatial layout conditioning for text-to-image diffusion models}

\author{Guillaume Couairon$^*$\\
Meta AI, Sorbonne Universit\'e\\
\and
Marl\`ene Careil$^*$\\
Meta AI\\
LTCI, T\'el\'ecom Paris, IP Paris\\               
\and
Matthieu Cord\\
Sorbonne Universit\'e, Valeo.ai
\and
St\'ephane Lathuili\`ere\\
LTCI, T\'el\'ecom Paris, IP Paris\\
\and
Jakob Verbeek\\
Meta AI
}

\maketitle
\ificcvfinal\thispagestyle{empty}\fi

\def\thefootnote{*}\footnotetext{These authors contributed equally to this work.}\def\thefootnote{\arabic{footnote}}

\begin{abstract}
Large-scale text-to-image diffusion models have  significantly improved the state of the art in generative image modeling and allow for an intuitive and powerful user interface to drive the image generation process.
Expressing spatial constraints, e.g.\ to position specific objects in particular locations, is cumbersome using text; and current text-based image generation models are not able to accurately follow such instructions. 
In this paper we consider image generation from text associated with segments on the image canvas, which combines an intuitive natural language interface with  precise spatial control over the generated content.
We propose ZestGuide, a ``zero-shot'' segmentation guidance approach that can be plugged into pre-trained text-to-image diffusion models, and does not require any additional training. 
It leverages implicit segmentation maps that can be extracted from  cross-attention layers, and uses them to align the generation with  input masks.
Our experimental results  combine high image quality with accurate alignment of generated content with input segmentations, and improve over prior work both quantitatively and qualitatively, including methods that require training on images with corresponding segmentations. Compared to Paint with Words, the previous state-of-the art in image generation with zero-shot segmentation conditioning, we improve by 5 to 10 mIoU points on the COCO dataset with similar FID scores.
\end{abstract}
\section{Introduction}

The ability of diffusion models to generate high-quality images has garnered widespread attention from the research community as well as the general public.
Text-to-image models, in particular, have demonstrated astonishing capabilities when trained on vast web-scale datasets~\cite{gafni22arxiv,ramesh22dalle2,rombach21arxiv,saharia22nips}.
This has led to the development of numerous image editing tools that facilitate content creation and aid creative media design~\cite{hertz2022prompt,meng2022sdedit,saharia2022palette}.
Textual description is an intuitive and powerful manner to  condition image generation. With a simple text prompt, even non-expert users can accurately describe their desired image and easily obtain  corresponding results. 
A single text prompt can effectively convey information about the objects in the scene, their interactions, and the overall style of the image. 
Despite their versatility, text prompts may not be the optimal choice for achieving fine-grained spatial control. Accurately describing the pose, position, and shape of each object in a complex scene with words can be a cumbersome task. Moreover, recent works have shown the limitation of diffusion models to follow spatial guidance expressed in natural language~\cite{avrahami2022spatext,casanova22}. 

\begin{figure}
    \centering
    \includegraphics[width=\linewidth]{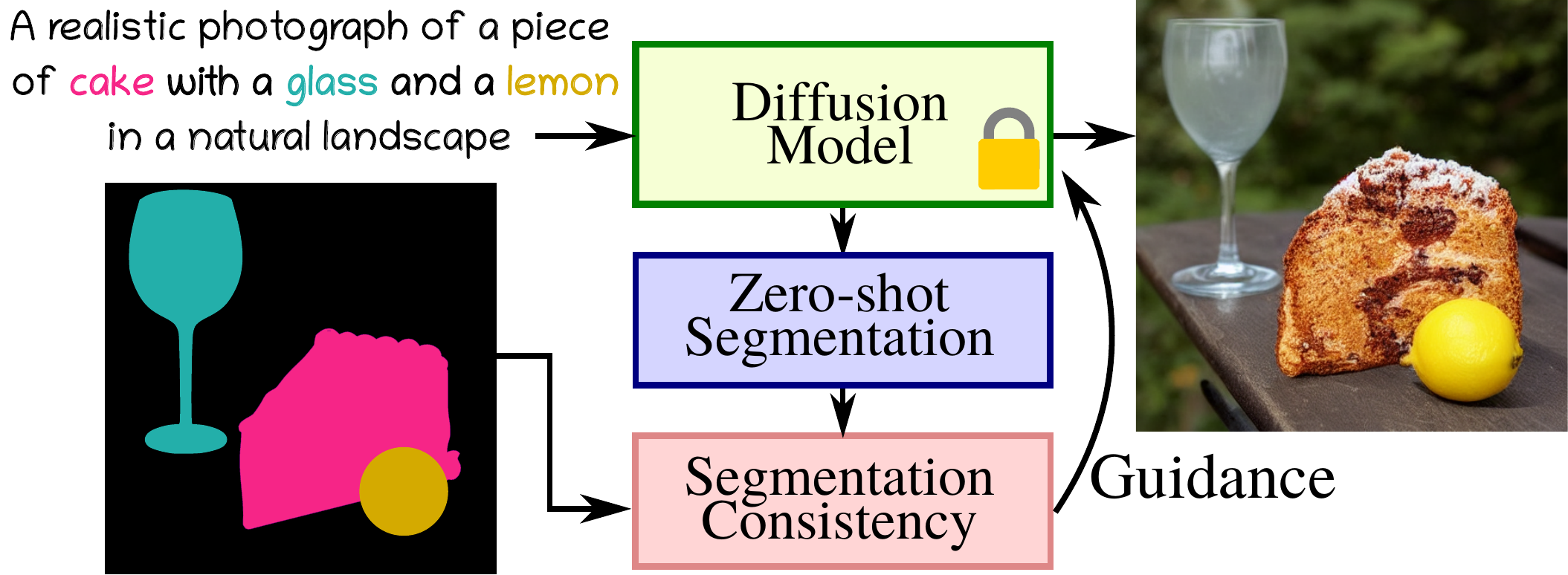}
    \myvspace
    \caption{In \ours the  image generation  is guided by the gradient of a loss  computed between the input segmentation and a segmentation recovered from attention in a text-to-image diffusion model. The approach does not require any additional training of the pretrained text-to-image diffusion model to solve this task.  
    }
    \label{fig:teaser}
\end{figure}

On the contrary, semantic image synthesis is a conditional image generation task that allows for detailed spatial control, by  providing a semantic map to indicate the desired class label for each pixel. 
Both adversarial \cite{park19cvpr1,sushko21iclr} and diffusion-based ~\cite{wang22arxiv,wang2022semantic} approaches have been  explored to generate high-quality and diverse images. However, these approaches rely heavily on large datasets with tens to hundreds of thousands of images annotated with pixel-precise label maps, which are  expensive to acquire and inherently limited in the number of class labels.

Addressing this issue, Balaji \etal~\cite{balaji22} showed that semantic image synthesis can be achieved using a pretrained text-to-image diffusion model in a zero-shot manner. Their training-free approach modifies the attention maps in the cross-attention layers of the diffusion model, allowing both spatial control and natural language conditioning. 
Users can input a text prompt along with a segmentation map that indicates the spatial location corresponding to parts of the caption. 
Despite their remarkable quality, the generated images tend to only roughly align with the input segmentation map.

\begin{figure}
 \def\myim#1{\includegraphics[width=27mm,height=27mm]{#1}}
 \footnotesize
     \centering
   \setlength\tabcolsep{1pt}
   \renewcommand{\arraystretch}{0.2}
     \begin{tabular}{ccc}
       ``A \yels{cat} wearing       &`` A \reds{dog} looking   &  ``\reds{Astronauts} on the  \\
       a \reds{dress}.'' &at the sunrise& street with rainbow \\
       & behind the \blues{fuji}.''& in outer space '' \\
 \myim{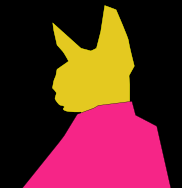} & 
\myim{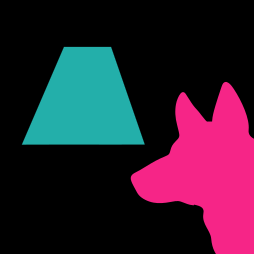} & 
\myim{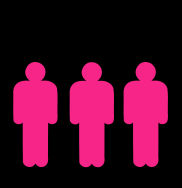}\\
 \myim{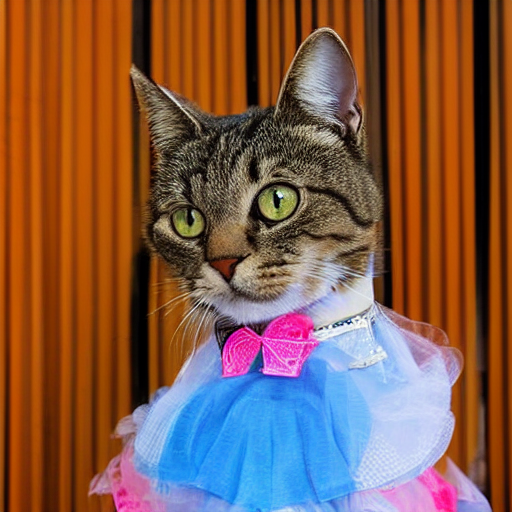} & 
\myim{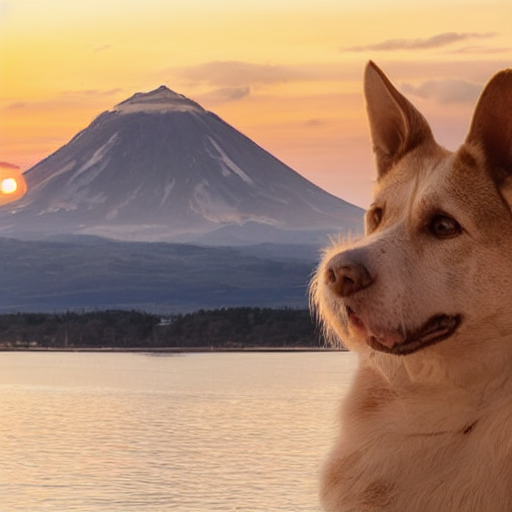} & 
\myim{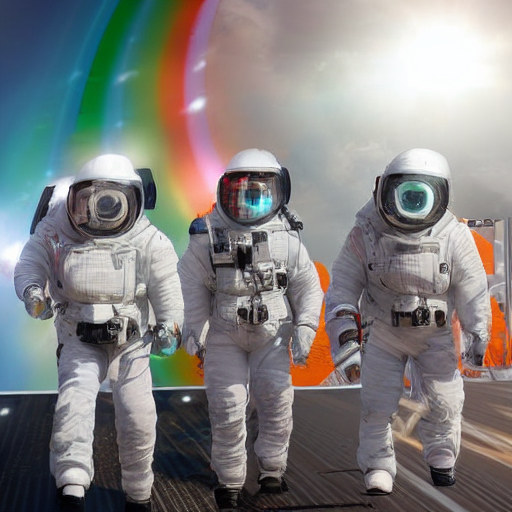}

\end{tabular}
\myvspace
\caption{\ours generates images conditioned on segmentation maps with corresponding free-form textual descriptions.}
\label{fig:quali}
\end{figure}

To overcome this limitation, we propose a novel approach called \ours, 
 short for  ZEro-shot SegmenTation GUIDancE, which empowers a pretrained text-to-image diffusion model to enable image generation conditioned on segmentation maps with corresponding free-form textual descriptions,  
 see examples presented in \fig{quali}.
\ours is designed to produce images which more accurately adhere to the conditioning semantic map.
Our zero-shot approach builds upon classifier-guidance techniques that allow for conditional generation from a pretrained unconditional diffusion model~\cite{dhariwal21nips}. 
These techniques utilize an external classifier to steer the iterative denoising process of diffusion models toward the generation of an image corresponding to the condition. 
While these approaches have been successfully applied to various forms of conditioning, such as class labels~\cite{dhariwal21nips} and semantic maps~\cite{bansal2023universal}, they still rely on pretrained recognition models. 
In the case of semantic image synthesis, this means that an image-segmentation network must be trained, which (i) violates our zero-shot objective, and (ii) allows each segment only to be conditioned on a single class label.
To circumvent the need for an external classifier, our approach takes advantage of the spatial information embedded in the cross-attention layers of the diffusion model to achieve zero-shot image segmentation. 
Guidance is then achieved by comparing a segmentation extracted from the attention layers with the conditioning map, eliminating the need for an external segmentation network.
In particular, \ours  computes a loss between the inferred segmentation and the input segmentation, and uses the gradient of this loss to guide the noise estimation process, allowing conditioning on free-form text rather than just class labels. Our approach does not require any training or fine-tuning on top of the text-to-image model. 

We conduct extensive experiments and compare our \ours to various approaches  introduced in the recent literature. 
Our results demonstrate state-of-the-art performance, %
improving both quantitatively and qualitatively over prior approaches.
Compared to Paint with Words, the previous state-of-the art in image generation with zero-shot segmentation conditioning, we improve by 5 to 10 mIoU points on the COCO dataset with similar FID scores.

In summary, our contributions are the following:
\begin{itemize}[noitemsep,topsep=0pt]
\item We introduce \ours, a zero-shot method for image generation from segments with text, designed to achieve high accuracy with respect to the conditioning map. We employ the attention maps of the cross-attention layer to perform zero-shot segmentation allowing classifier-guidance without the use of an external classifier.
\item We obtain excellent experimental results, improving over existing both zero-shot and training-based approaches both quantitatively and qualitatively.
\end{itemize}

\section{Related work}

\mypar{Spatially conditioned generative image models} 
Following seminal works on image-to-image translation~\cite{isola17cvpr}, spatially constrained image generation has been extensively studied.
In particular, the task of semantic image synthesis  consists in generating images conditioned on masks where each pixel is annotated with a class label. 
Until recently, GAN-based approaches were prominent with methods such as SPADE~\cite{park19cvpr1}, and OASIS~\cite{sushko21iclr}. 
Alternatively, autoregressive  transformer models over discrete VQ-VAE~\cite{oord17nips} representations to synthesize images from text and semantic segmentation maps have been considered~\cite{esser21cvpr,gafni22arxiv,razavi19nips}, as well as non-autoregressive models with faster sampling~\cite{chang22cvpr, lezama22eccv}.

Diffusion models have recently emerged as a very powerful class of generative image models, and have also been explored for semantic image synthesis. 
For example, PITI~\cite{wang22arxiv} finetunes GLIDE~\cite{nichol21arxiv}, a large pretrained text-to-image generative model, by replacing its text encoder with an encoder of semantic segmentation maps. 
SDM~\cite{wang2022semantic} trains a diffusion model using SPADE blocks to condition the denoising U-Net on the input segmentation. 

The iterative nature of the decoding process in diffusion models,  allows so called ``guidance'' techniques to strengthen the input conditioning during the decoding process.
For example,  \textit{classifier guidance}~\cite{dhariwal21nips} has been used for class-conditional image generation by applying a pretrained classifier on the partially decoded image, and using the gradient of the classifier to guide the generation process to output an image of the desired class. 
It has since been extended to take as input other constraints such as for the tasks of inpainting, colorization, and super-resolution~\cite{saharia2022palette}. 
For semantic image synthesis, the gradient of a pretrained semantic segmentation network can be used as guidance~\cite{bansal2023universal}.
This approach, however, suffers from two drawbacks. 
First, only the classes recognized by the segmentation model can be used to constrain the image generation, although this can to some extent be alleviated   using an open-vocabulary segmentation model like CLIPSeg~\cite{luddecke22cvpr}. 
The second drawback is that this approach requires a full forwards-backwards pass through the external segmentation network in order to obtain the gradient at each step of the diffusion process, which requires additional memory and compute on top of the diffusion model itself. 

 While there is a vast literature on semantic image synthesis, it is more limited when it comes to the more general task of synthesizing images conditioned on masks with free-form textual descriptions.
 SpaText~\cite{avrahami2022spatext} finetunes a large pretrained text-to-image diffusion model with an additional input of segments annotated with free-form texts. 
 This representation is extracted from a pretrained multi-modal CLIP encoder~\cite{radford21clip}: using visual  embeddings during training, and swapping to textual  embeddings during inference. 
 GLIGEN~\cite{li2023gligen} adds trainable layers on top of a pretrained diffusion models to extend conditioning from text to bounding boxes and pose.  
 These layers take the form of additional attention layers that incorporate the local information. 
 T2I~\cite{mou2023t2i} and ControlNet~\cite{zhang2023adding} propose to extend a pretrained and frozen diffusion model with  small adapters for task-specific spatial control using pose, sketches, or segmentation maps.
 All these methods require to be trained on a large dataset with segmentation annotations, which is computationally costly and requires specialized training data.

\mypar{Train-free adaptation of text-to-image diffusion models}
Several recent studies~\cite{chefer2023attend,feng2022training,hertz2022prompt,  parmar2023zero} found that the positioning content in generated images from large text-to-image diffusion models correlates with the cross-attention maps, which diffusion models use to condition the denoising process on the conditioning text. 
This correlation can be leveraged to adapt text-to-image diffusion  at inference time for various downstream applications. 
For example, \cite{chefer2023attend,feng2022training} aim to achieve better image composition and attribute binding. 
Feng \etal~\cite{feng2022training} design a pipeline to associate attributes to objects and incorporate this linguistic structure by modifying values in cross-attention maps. 
Chefer \etal~\cite{chefer2023attend} guide the generation process with gradients from a loss aiming at strengthening attention maps activations of ignored objects. 

Zero-shot image editing was explored in several works~\cite{couairon22,hertz2022prompt,meng2022sdedit,parmar2023zero}. 
SDEdit~\cite{meng2022sdedit} consists in adding noise to an input image, and  denoising it to project it  to the  manifold of natural images. It is mostly applied on transforming sketches into natural images. 
Different from SDEdit, in which there is no constraint on which part of the image to modify, DiffEdit~\cite{couairon22} proposes a method to automatically find masks corresponding to where images should be edited for a given prompt modification. 
Prompt-to-Prompt~\cite{hertz2022prompt} and pix2pix-zero~\cite{parmar2023zero} act on cross-attention layers  by manipulating 
attention  layers and imposing a struture-preserving loss on the attention maps, respectively. 

Closer to our work, eDiff-I~\cite{balaji22} proposes a procedure to synthesize images from segmentation maps with local free-form texts. They do so by rescaling attention maps at locations specified by the input semantic masks, similarly to \cite{ma2023directed} for controlling the position of objects.
MultiDiffusion~\cite{bartal23} fuses multiple generation processes constrained by shared parameters from a pretrained diffusion model by solving an optimization problem, and applying it to panorama generation and spatial image guidance.
In~\cite{bansal2023universal}  a pretrained segmentation net  guides image generation to respect a segmentation  map during the denoising process. %

\section{Method}

\begin{figure}
    \centering
    \includegraphics[width=0.9\linewidth]{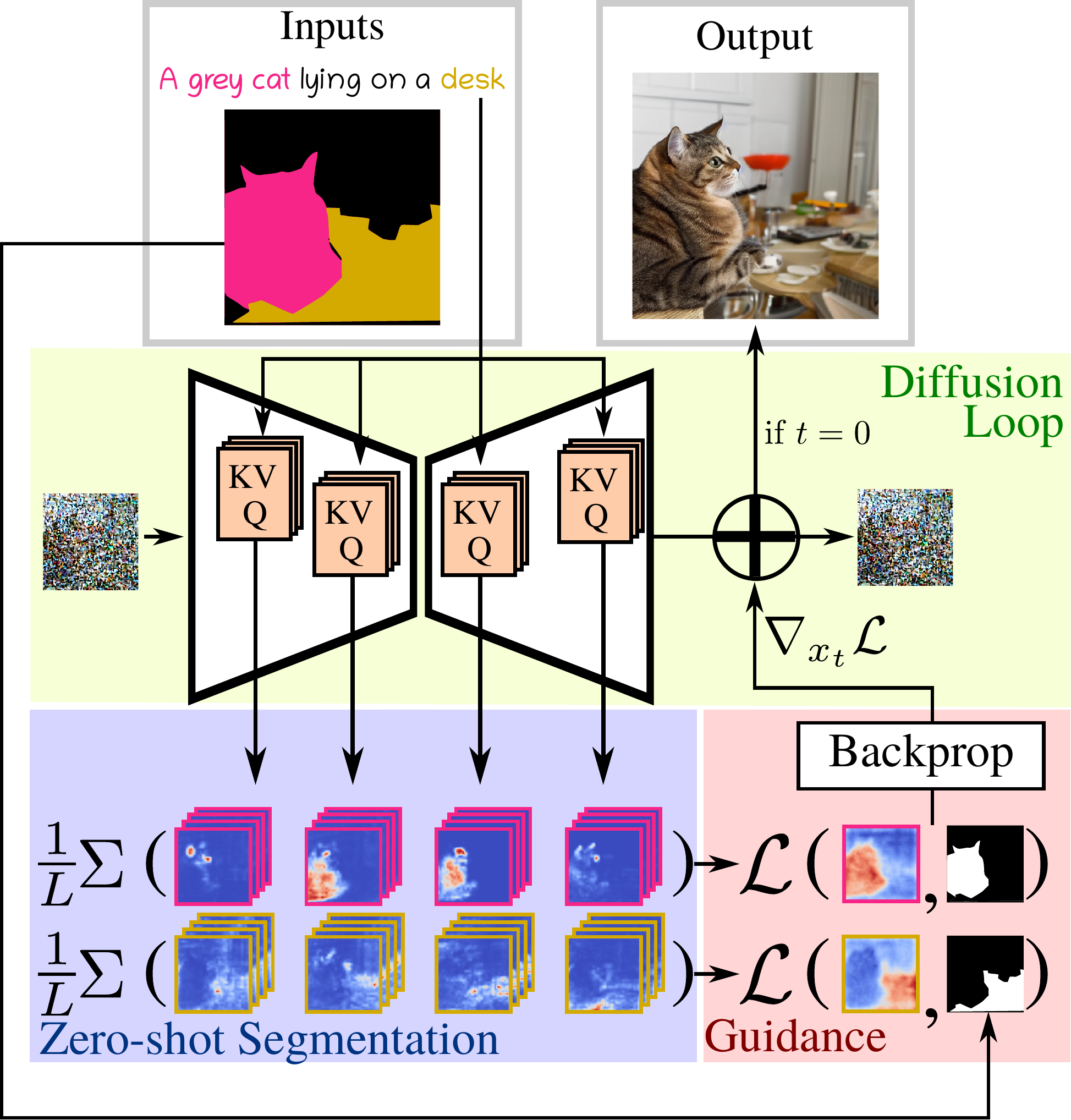}
\myvspace
    \caption{ \ours extracts segmentation maps from text-attention layers in pretrained diffusion models, and uses them to align the generation with input masks via gradient-based guidance.     }
    \label{fig:method}
\end{figure}

In this section, we provide a concise introduction  of diffusion models in \sect{motiv} before presenting our novel approach, \ours, which extends pretrained text-to-image diffusion models to enable conditional generation of images based on segmentation maps and associated text without requiring additional training, as described in \sect{zeroshot}. 
In \fig{method} we provide an overview of \ours.

\subsection{Preliminaries}  
\label{sec:motiv}

\mypar{Diffusion models} 
Diffusion models~\cite{ho20neurips}  approximate a data distribution by gradually denoising a random variable drawn from a  unit Gaussian prior. 
The denoising function is trained to invert a diffusion process, which maps sample $\xmat_0$ from the data distribution to the prior by sequentially adding a small Gaussian noise for a large number of timesteps $T$. 
In practice, a noise estimator neural network $\epsilon_\theta(\xmat_t,t)$ is trained to denoise inputs $\xmat_t = \sqrt{\alpha_t} \xmat_0 + \sqrt{1-\alpha_t} \epsilon$, which are data points $\xmat_0$ corrupted with Gaussian noise $\epsilon$ where $\alpha_t$ controls the level of noise, from $\alpha_0 = 1$ (no noise) to $\alpha_T \simeq 0$ (pure noise). Given the trained noise estimator,  samples from the model can be drawn by sampling Gaussian noise $\xmat_T \sim \gaussian$, and iteratively applying the denoising Diffusion Implicit Models (DDIM) equation~\cite{song2020denoising}.

Rather than applying diffusion models directly in pixel space, it is more efficient to apply them in the latent space of a learned autoencoder~\cite{rombach21arxiv}.

Text-conditional generation can be achieved by  providing an encoding $\rho(y)$ of the text $y$ as additional input to the noise estimator $\epsilon_\theta(\xmat_t,t,\rho(y))$ during training. 
The noise estimator $\epsilon_\theta$ is commonly implemented using the U-Net architecture, and 
the text encoding takes the form of a sequence of token embeddings obtained using a transformer model.
This sequence is usually processed with cross-attention layers in the U-Net, where keys and values are estimated from the text embedding.

\mypar{Classifier guidance} Classifier guidance is a technique  for conditional sampling of diffusion models~\cite{sohl-dickstein15icml,song21iclr2}. 
Given a label $c$  of an image $\xmat_0$, samples from the posterior distribution $p(\xmat_0 | c)$ can be obtained by sampling each transition in the generative process according to $p(\xmat_t | \xmat_{t+1}, c) \propto p(\xmat_t | \xmat_{t+1}) p(c | \xmat_t)$ instead of $p(\xmat_t | \xmat_{t+1})$. Dhariwal and Nichol~\cite{dhariwal21nips} show that DDIM sampling can be extended to sample the posterior distribution, with the following modification for the noise estimator $\epsilon_\theta$:
\begin{equation}
\label{eq:cg}
\begin{split}
\tilde{\epsilon}_\theta(\xmat_t, t, \rho(y)) &= \epsilon_\theta(\xmat_t, t, \rho(y)) \\ &- \sqrt{1 - \alpha_t} \nabla_{\xmat_t} p(c | \xmat_t).
\end{split}
\end{equation}
Classifier guidance can be straightforwardly adapted to generate images conditioned on semantic segmentation maps by replacing the classifier by a segmentation network which outputs a label distribution for each pixel in the input image.
However this approach suffers from several weaknesses: (i) it requires to train an external  segmentation model; (ii) semantic synthesis is bounded to the set of classes modeled by the segmentation model; (iii) it is computationally expensive since it implies back-propagation through both the latent space decoder and the segmentation network at every denoising step. 
To address these issues, we propose to employ the cross-attention maps computed in the denoising model $\epsilon_\theta$ of text-to-image diffusion models to achieve zero-shot segmentation. This has two major advantages: first, there is no need to decode the image at each denoising step; second, our zero-shot segmentation process is extremely lightweight, so the additional computational cost almost entirely comes from backpropagation through the U-Net, which is a relatively low-cost method for incorporating classifier guidance.

\begin{figure}

\centering
\small
\setlength{\tabcolsep}{1pt}

\def\myim#1{\includegraphics[width=20mm,height=20mm]{figs/attns_woGuid/#1.png}}

\begin{tabular}{cccc} 
 \myim{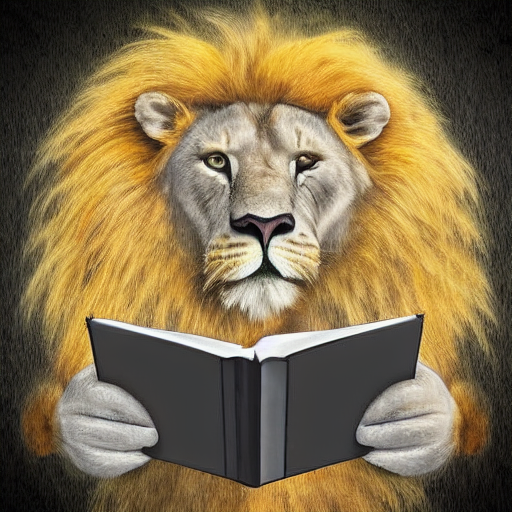} & \myim{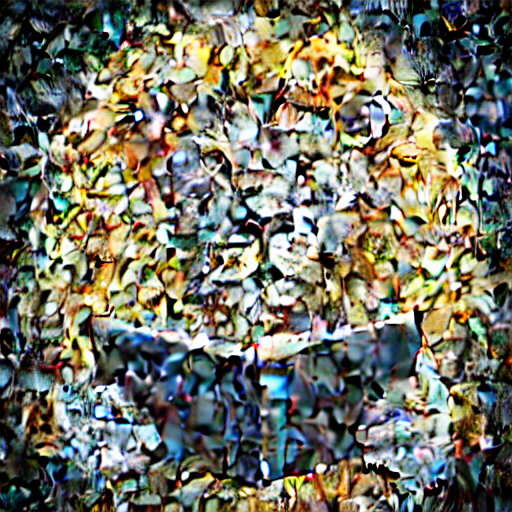} & \myim{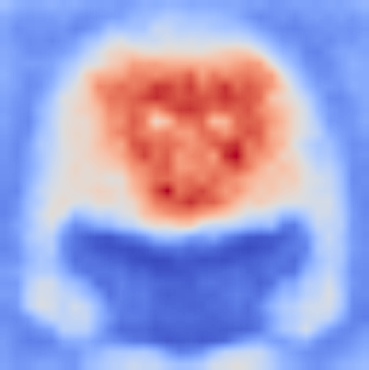} & \myim{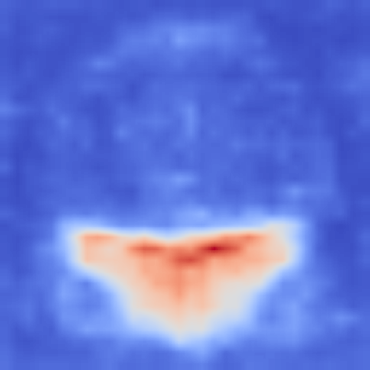}  \\
Generated & U-Net input & Lion attn. & Book attn.
\end{tabular}

\def\myim#1{\includegraphics[width=12mm]{figs/attns_woGuid/#1.png}}

\begin{tabular}{ccccccc} 
\begin{sideways}Lion\end{sideways}& 
\myim{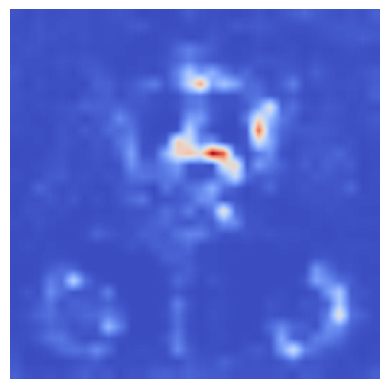} & \myim{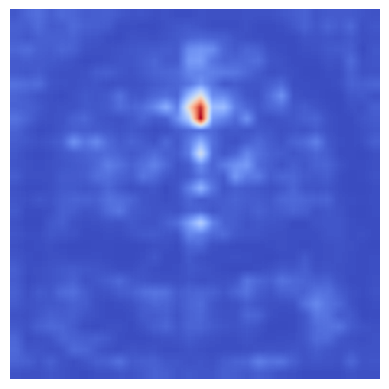} & \myim{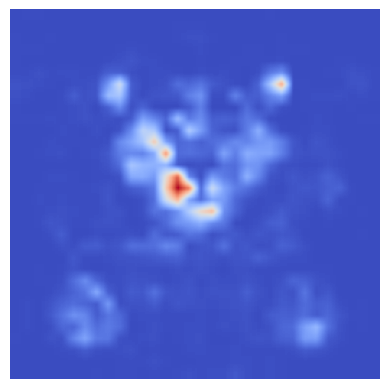} & \myim{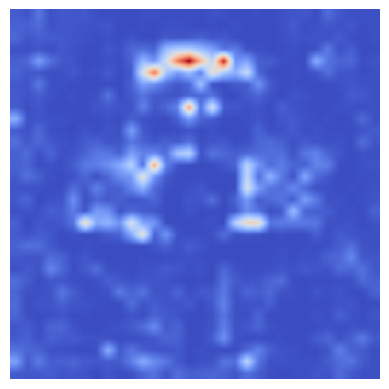} & \myim{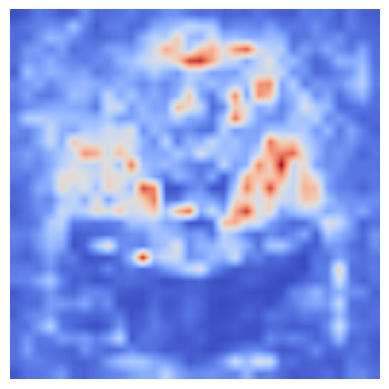} & \myim{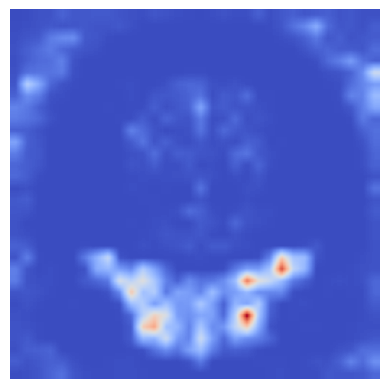} \\
\begin{sideways}Book\end{sideways}& 
\myim{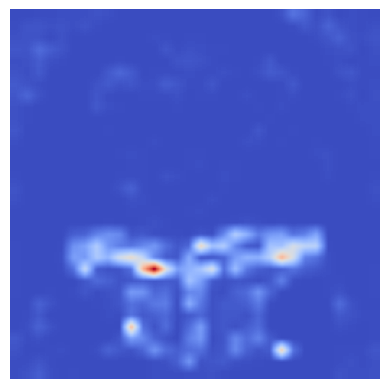} & \myim{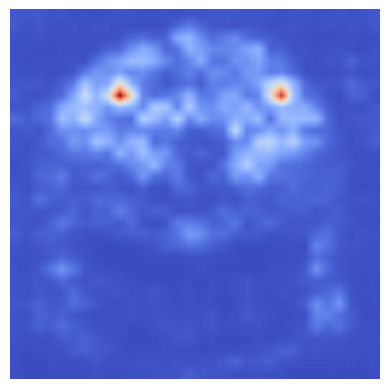} & \myim{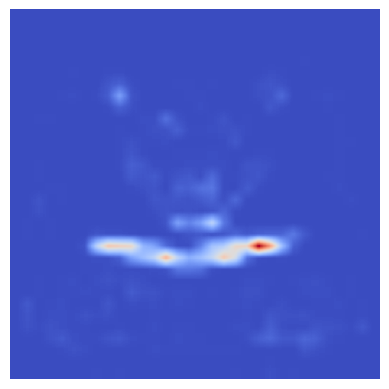} & \myim{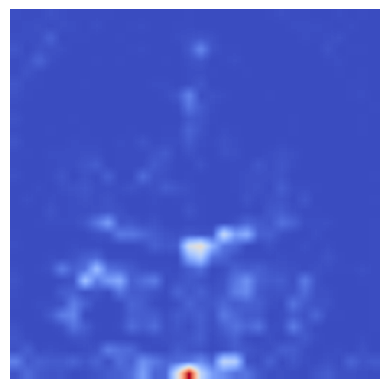} & \myim{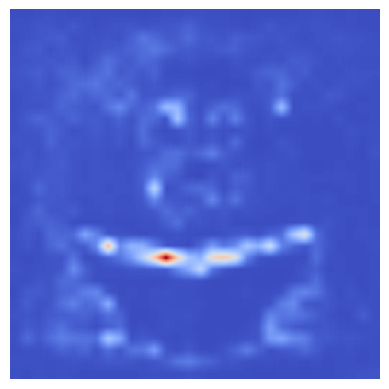} & \myim{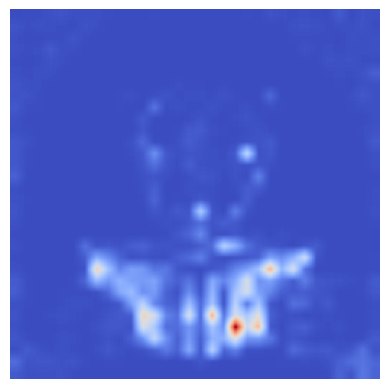} 
\end{tabular}
\vspace{1mm}
\caption{
Top, from left to right: image generated from the prompt \textit{``A lion reading a book.''}, the noisy input to the U-Net at $t=20$, cross-attention averaged over different heads and U-Net layers for ``Lion'' and ``Book''.
Bottom: individual attention heads.
}
\label{fig:attention}
\end{figure}

\subsection{Zero-shot segmentation with attention}
\label{sec:zeroshot}

To condition the image generation, we consider a text prompt of length $N$ denoted as $\mathcal{T}=\{T_1, \dots, T_N\}$, and a set of $K$ binary segmentation maps $\Smat = \{\Smat_1,\dots,\Smat_K\}$.
Each segment $\Smat_i$ is associated with a subset $\mathcal{T}_i\subset\mathcal{T}$.

\mypar{Attention map extraction} 
 We leverage cross-attention layers  of the diffusion U-Net to segment the image as it is  generated. 
The attention maps are computed independently for every layer and head in the U-Net. 
For layer $l$, the queries $\Qmat_l$ are computed from local image features using a linear projection layer. %
Similarly, the keys $\Kmat_l$ are computed from the word descriptors $\mathcal{T}$ with another layer-specific linear projection. %
The cross-attention from  image features to text tokens, is  computed as 
\begin{equation}
\Amat_l = \textrm{Softmax}\left(\frac{\Qmat_l \Kmat_l^T}{\sqrt{d}}\right),
\end{equation}
where the query/key  dimension $d$ is used to normalize the softmax energies~\cite{vaswani17nips}.
Let  $\Amat^n_l = \Amat_l[n]$ denote the attention of image features \wrt specific text token $T_n\in\mathcal{T}$ in layer $l$ of the U-Net. 
To simplify notation, we use $l$  to index over both the layers of the U-Net as well as the  different attention heads in each layer.
In practice, we find that the attention maps provide meaningful localisation information, but only when they are averaged across different attention heads and feature layers. 
See \fig{attention} for an illustration.

Since the attention maps have varying resolutions depending on the layer, we upsample them to the highest resolution.
Then, for each segment  we compute an  attention map $\Smat_i$ by averaging attention maps across layers and text tokens associated with the segment:
\begin{equation}
\label{eq:att}
\hat{\Smat}_i = \frac{1}{L} \sum_{l=1}^L \sum_{j=1}^N \; \llbracket T_j\in \mathcal{T}_i \rrbracket\;\Amat^{l}_j,
\end{equation}
where $\llbracket\cdot\rrbracket$ is the Iverson bracket notation which is one if the argument is true and zero otherwise.

\mypar{Spatial self-guidance} 
We compare the averaged attention maps to the input segmentation using a sum of binary cross-entropy losses computed separately for each segment:
\begin{equation}
\mathcal{L}_\textrm{Zest} = \sum_{i=1}^K \bigg( \mathcal{L}_\textrm{BCE}(\hat{\Smat}_i, \Smat_i) + \mathcal{L}_\textrm{BCE}(\frac{\hat{\Smat}_i}{\Vert \hat{\Smat}_i \Vert_\infty}, \Smat_i) \bigg).
\label{eq:segmentationloss}
\end{equation}
In the second loss term, we normalized the attention maps $\hat{\Smat}_i$ independently for each object. %
This choice is motivated by two observations. Firstly, we found that averaging softmax outputs across heads, as described in Eq. \eqref{eq:att}, generally results in low maximum values in $\hat{\Smat}_i$. By normalizing the attention maps, we make them more comparable with the conditioning $\Smat$. Secondly, we observed that estimated masks can have different maximum values across different segments resulting in varying impacts on the overall loss. Normalization helps to balance the impact of each object. However, relying solely on the normalized term is insufficient, as the normalization process cancels out the gradient corresponding to the maximum values.

We then use DDIM sampling with classifier guidance based on the gradient of this loss. We use \Eq{cg} to compute the modified noise estimator at each denoising step. Interestingly, since $\xmat_{t-1}$ is computed from $\tilde{\epsilon}_\theta(\xmat_t)$, this conditional DDIM sampling corresponds to an alternation of regular DDIM updates and gradient descent updates on $\xmat_t$ of the loss $\mathcal{L}$, with a fixed learning rate $\eta$ multiplied by a function $\lambda(t)$ monotonically decreasing from one to zero throughout the generative process. In this formulation, the gradient descent update writes: 

\begin{equation}
\tilde{\xmat}_{t-1}  = \xmat_{t-1} - \eta \cdot \lambda(t) \frac{\nabla_{\xmat_t} \mathcal{L}_\textrm{Zest}}{\Vert \nabla_{\xmat_t} \mathcal{L}_\textrm{Zest} \Vert_\infty}. 
\label{eq:ZestUpdate}
\end{equation}
Note that differently from \Eq{cg}, the gradient is normalized to make updates more uniform in strength across images and denoising steps. We note that the learning rate $\eta$ can be set freely, which, as noted by \cite{dhariwal21nips}, corresponds to using a renormalized classifier distribution in classifier guidance. As in \cite{balaji22}, we define a hyperparameter $\tau$ as the fraction of steps during which classifier guidance is applied. Preliminary experiments suggested that classifier guidance is only useful in the first $50\%$ of DDIM steps, and we set $\tau = 0.5$ as our default value, see \sect{ablations} for more details.

\section{Experiments}

We present our experimental setup in \sect{setup}, followed by our main results in \sect{results} and ablations in \sect{ablations}.

\subsection{Experimental setup}
\label{sec:setup}

\mypar{Evaluation protocol} 
We use the COCO-Stuff  validation split, which contains  5k  images annotated with fine-grained pixel-level segmentation masks across 171 classes, and five  captions describing each image~\cite{caesar18cvpr}.
We adopt three different setups to evaluate our approach and  to compare to baselines.  
In all three settings, the generative diffusion model is  conditioned on one of the five captions corresponding to the segmentation map, but they differ in  the segmentation maps used for spatial conditioning.

\begin{table*}
\centering
 \setlength{\tabcolsep}{3pt} 
  {\small
\begin{tabular}{lcccccccccccc}
\toprule
\multirow{2}{*}{\textbf{Method}}  & \textbf{Free-form}  & \textbf{Zero-} &    \multicolumn{3}{c}{\textbf{Eval-all}} &   \multicolumn{3}{c}{\textbf{Eval-filtered}} & \multicolumn{3}{c}{\textbf{Eval-few}} \\  
  & \textbf{mask texts} & \textbf{shot} &    $\downarrow$\textbf{FID}  &    $\uparrow$\textbf{mIoU}   & $\uparrow$\textbf{CLIP}   &    $\downarrow$\textbf{FID}  &    $\uparrow$\textbf{mIoU} &    $\uparrow$\textbf{CLIP}   &$\downarrow$\textbf{FID}  &    $\uparrow$\textbf{mIoU} &   $\uparrow$\textbf{CLIP}  \\  
 \midrule

OASIS~\cite{sushko21iclr} & \xmark & \xmark & 15.0 & 52.1& --- &  18.2 & 53.7& --- & 46.8 & 41.4 & ---\\
SDM~\cite{wang2022semantic}  &  \xmark & \xmark & 17.2 & 49.3 & --- & 28.6 & 41.7 &  --- & 65.3 & 29.3 &  --- \\
SD w/ T2I-Adapter~\cite{mou2023t2i} &  \xmark & \xmark &  17.2 &  33.3  &  31.5 & 17.8 &  35.1  &  31.3 &  19.2  & 31.6  & 30.6 \\
 LDM w/ External Classifier  &  \xmark & \xmark  & 24.1 & 14.2 & 30.6 & 23.2 & 17.1 & 30.2 & 23.7 & 20.5 & 30.1\\

\midrule
 SD w/ SpaText~\cite{avrahami2022spatext}  &  \cmark & \xmark & \bf 19.8 & 16.8& 30.0& \bf 18.9 &19.2 &30.1& \bf 16.2 &23.8 &30.2\\

SD w/ PwW~\cite{balaji22} &  \cmark & \cmark  & 36.2 &21.2&29.4& 35.0& 23.5&29.5& 25.8 & 23.8 & 29.6 \\

LDM w/ MultiDiffusion\cite{bartal23} &  \cmark & \cmark  & 59.9 & 15.8 &23.9 &46.7 &18.6&25.8&21.1 & 19.6 & 29.0 \\
LDM w/ PwW  &  \cmark & \cmark  & 22.9 & 27.9 & 31.5&23.4 & 31.8 & \bf 31.4 &20.3 & 36.3 &  \bf 31.2  \\
LDM w/ \ours (ours)  &  \cmark & \cmark  & 22.8 & \bf 33.1 & \bf 31.9 &23.1 & \bf 43.3 & 31.3 & 21.0 & \bf 46.9 & 30.3\\
\bottomrule
\end{tabular}
}
\myvspace
\caption{Comparison  of \ours to  other methods in our three  evaluation settings. 
OASIS and SDM are trained from scratch on COCO, other methods are based on  pre-trained text-to-image models:  StableDiffusion (SD) or our latent diffusion model (LDM). 
Methods that do not allow for free-form text description of segments  are listed in the upper part of the table. 
Best scores in each part of the table are marked in bold.
For OASIS and SDM  the CLIP score is omitted as it is  not meaningful for methods that  don't condition on  text prompts.
}
\vspace{-3mm}
\label{tab:evalAllCls}
\end{table*}

The first evaluation setting, \emph{Eval-all}, conditions image generation on complete segmentation maps across all classes, similar to the evaluation setup in OASIS~\cite{sushko21iclr} and SDM~\cite{wang2022semantic}.
In the \emph{Eval-filtered} setting, segmentation maps are modified by removing all segments occupying less than $5\%$ of the image, which is more representative of real-world scenarios where users may not provide segmentation masks for very small objects.
Finally, in \emph{Eval-few} we retain between one and three segments, each covering at least $5\%$ of the image, similar to the setups in~\cite{avrahami2022spatext,bartal23}. 
It is the most realistic setting, as users may be interested in  drawing only a few objects, and therefore the focus of our evaluation. Regarding the construction of the text prompts, we follow~\cite{avrahami2022spatext} and concatenate the annotated prompt of COCO with the list of class names corresponding to the input segments.

\mypar{Evaluation metrics} 
We use the two standard metrics to evaluate semantic image synthesis, see \eg~\cite{park19cvpr1,sushko21iclr}.  
 Fr\'echet Inception Distance (FID)~\cite{heusel17nips} captures both image quality and diversity. We compute FID with InceptionV3 and generate 5k images. The reference set is the original COCO validation set, and we use code from~\cite{parmar2021cleanfid}.
The mean Intersection over Union (mIoU) metric  measures to what extent the generated images respect the spatial conditioning. We additionally compute a CLIP score that measures alignment between captions and generated images. All methods, including ours, generate images at resolution $512\times 512$, except OASIS and SDM, for which we use available pretrained checkpoints synthesizing images at resolution $256\times 256$, which we upsample to $512\times 512$. %

\mypar{Baselines} 
We compare to baselines that are either trained from scratch, finetuned or training-free. 
The adversarial OASIS model~\cite{sushko21iclr} and diffusion-based SDM model~\cite{wang2022semantic} are both trained from scratch and conditioned on segmentation maps with classes of COCO-Stuff dataset. 
For SDM we use $T=50$ diffusion decoding steps.
T2I-Adapter~\cite{mou2023t2i} and SpaText~\cite{avrahami2022spatext} both fine-tune pre-trained text-to-image diffusion models  for spatially-conditioned image generation by incorporating additional trainable layers in the diffusion pipeline. 
Similar to Universal Guidance~\cite{bansal2023universal}, we  implemented a method in which we use classifier guidance based on the external pretrained segmentation network DeepLabV2 ~\cite{chen2017deeplab} to guide the generation process to respect a semantic map.
We also compare \ours to other zero-sot methods that adapt a pre-trained text-to-image diffusion model during inference. MultiDiffusion~\cite{bartal23} decomposes the denoising procedure into several diffusion processes, where each one focuses on one segment of the image and fuses all these different predictions at each denoising iteration.  
In~\cite{balaji22} a conditioning pipeline called ``\textit{paint-with-words}'' (PwW) is proposed, which manually modifies the values of attention maps. 
For a fair comparison, we evaluate these zero-shot methods on the same diffusion model used to implement our method. 
Note that SpaText, MultiDiffusion, PwW, and our method can be locally conditioned on free-form text, unlike Universal Guidance, OASIS, SDM and T2I-Adapter which can only condition on  COCO-Stuff classes.

\mypar{Text-to-image model} 
Due to concerns regarding the training data of Stable Diffusion~\cite{rombach21arxiv} (such as copyright infringements and consent), we refrain from experimenting with this model and  instead use a large diffusion model (2.2B parameters) trained on a proprietary dataset of 330M image-text pairs. 
We refer to this model as LDM. 
Similar to~\cite{rombach21arxiv} the model is trained on the latent space of an autoencoder, and we use an architecture for  the diffusion model based on GLIDE~\cite{nichol21arxiv}, with a T5 text encoder~\cite{raffel20jmlr}. 
With an FID score of 19.1 on the COCO-stuff dataset, our LDM model achieves image quality similar to that of Stable Diffusion, whose FID score was 19.0, while using an order of magnitude less training data.

\mypar{Implementation details}
For all experiments that use our LDM diffusion model, we use 50 steps of DDIM sampling with classifier-free guidance strength set to 3. For \ours results, unless otherwise specified, we use classifier guidance in combination with the PwW algorithm. We review this design choice in \sect{ablations}.

\begin{figure*}
 \def\myim#1{\includegraphics[width=27mm,height=27mm]{#1}}
 \def\myimCS#1{ \includegraphics[width=34.5mm,height=17.25mm]{samples/#1}}
 \footnotesize
     \centering
   \setlength\tabcolsep{0.5 pt}
   \renewcommand{\arraystretch}{0.2}
     \begin{tabular}{ccccccc}

 &``Two zebra standing  &``Five oranges  &  ``There is a dog  & ``A person  over a box'' & ``A train traveling & ``There is a woman \\
 &next to each other &with a red apple &  holding a Frisbee &  jumping a horse&  through rural country- & about to ski\\
 &in a dry grass field.'' & and a green apple.'' & in its mouth.''  & over a box'.'& side lined with trees.'' & down a hill.''\\
 &\myim{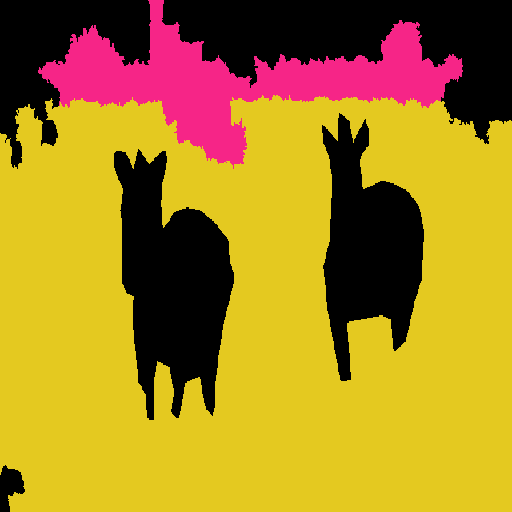} & 
\myim{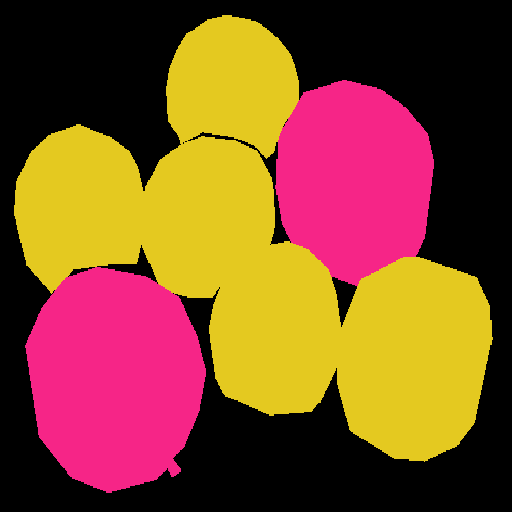} & 
\myim{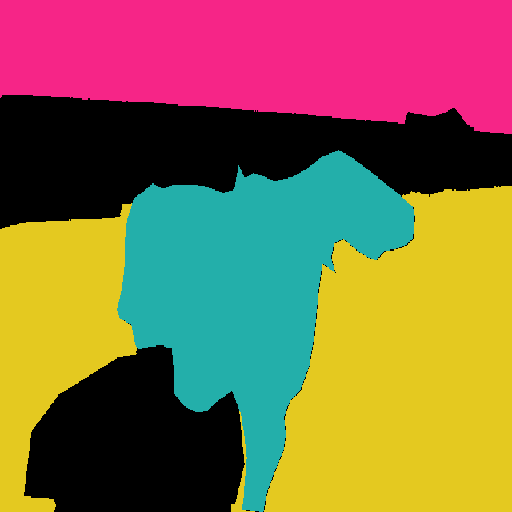} &  \myim{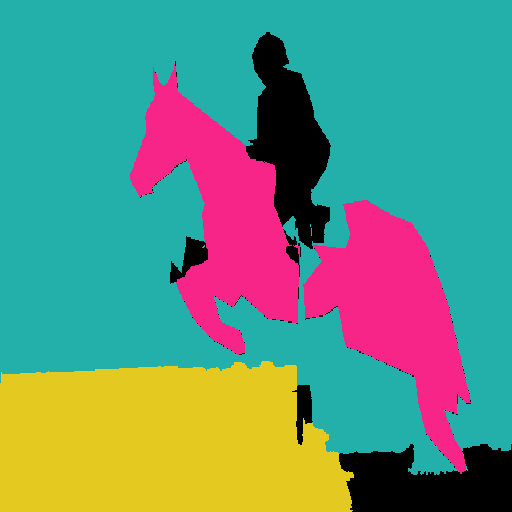}& \myim{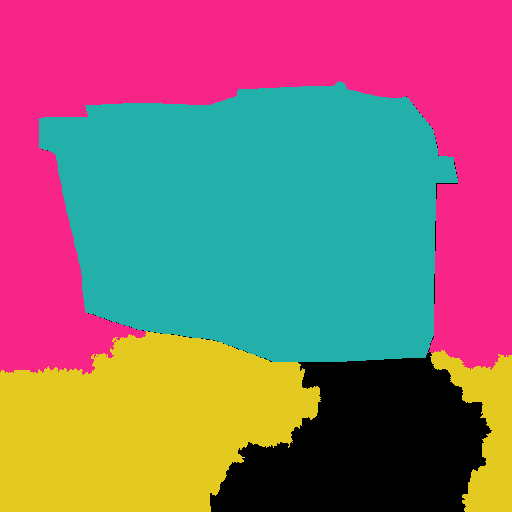} & \myim{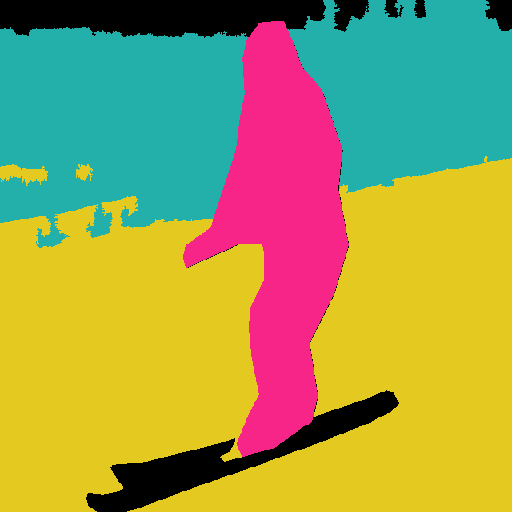} \\

\vspace{0.5mm} \\
& \red{plant} & \red{apple} &\red{sky}  & \red{horse} & \red{bush} & \red{person}\\
 &\yel{straw}& \yel{orange}& \yel{sand}  & \yel{fence} & \yel{grass} & \yel{snow}\\

 &&  & \blue{dog} &  \blue{tree} & \blue{train} & \blue{tree}\\
\begin{sideways}   \hspace{0.4cm}Ext. Classifier\end{sideways} &\myim{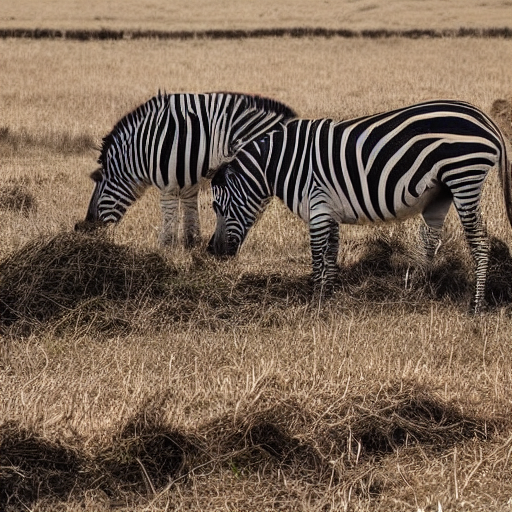}
&\myim{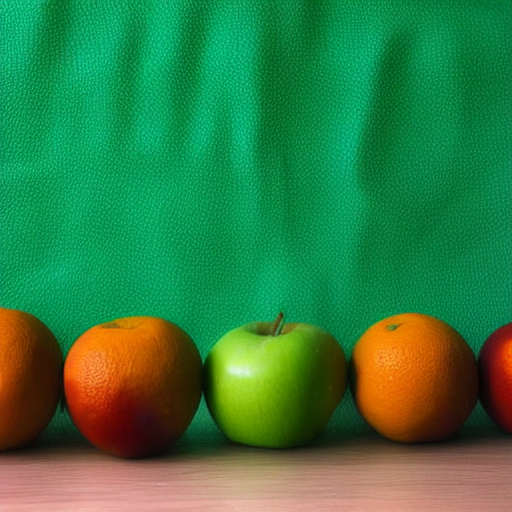}& 
\myim{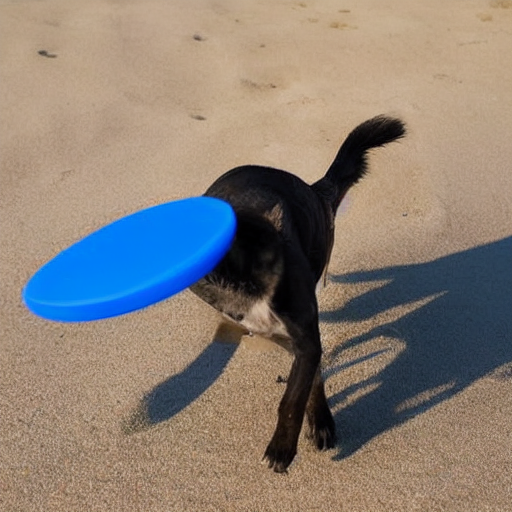}& \myim{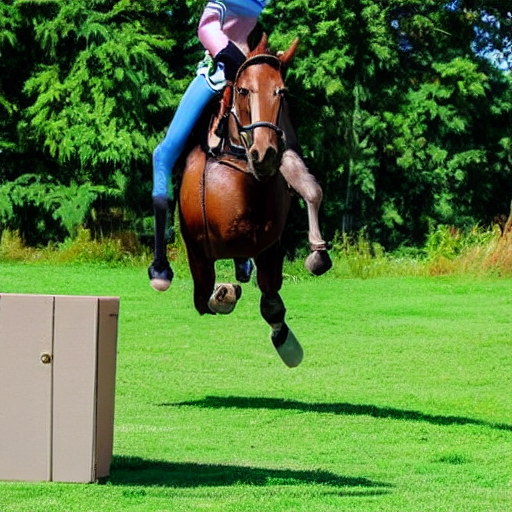}& \myim{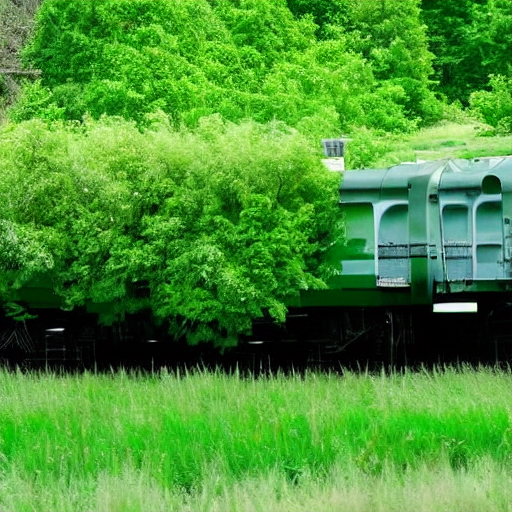}& \myim{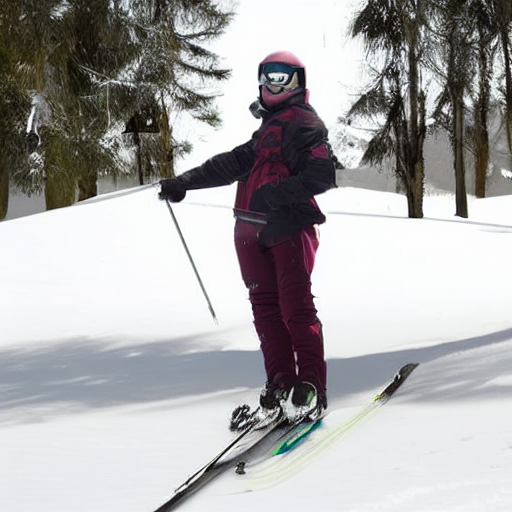} \\
\begin{sideways} \hspace{0.4cm}MultiDiffusion\end{sideways} &
\myim{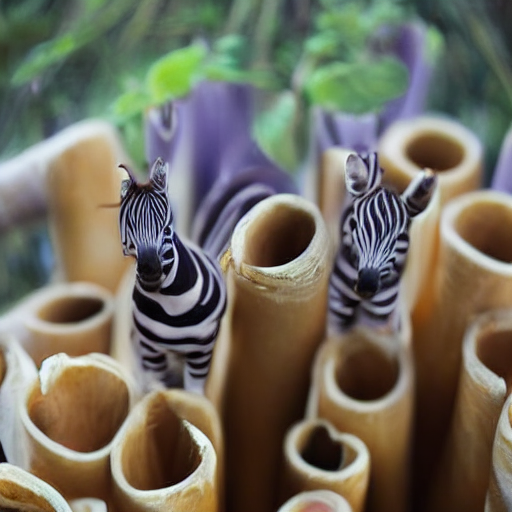}&
\myim{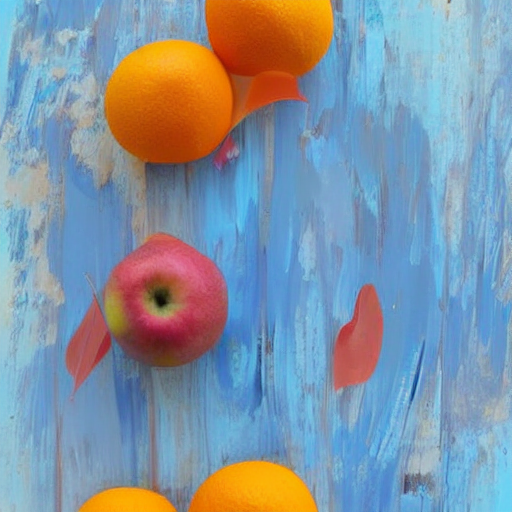}&\myim{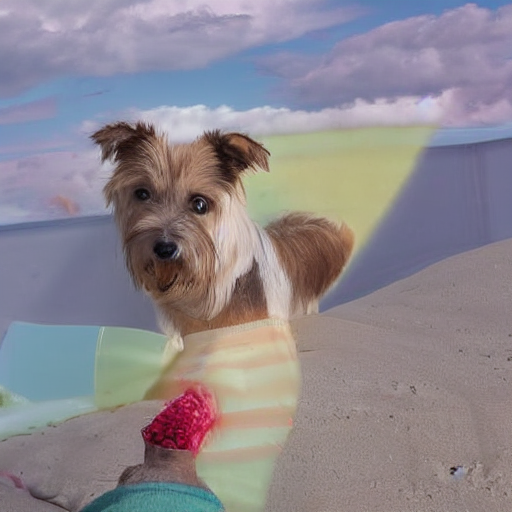}&\myim{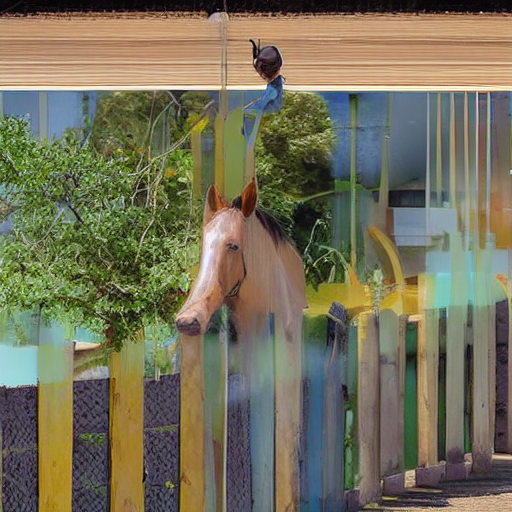}&\myim{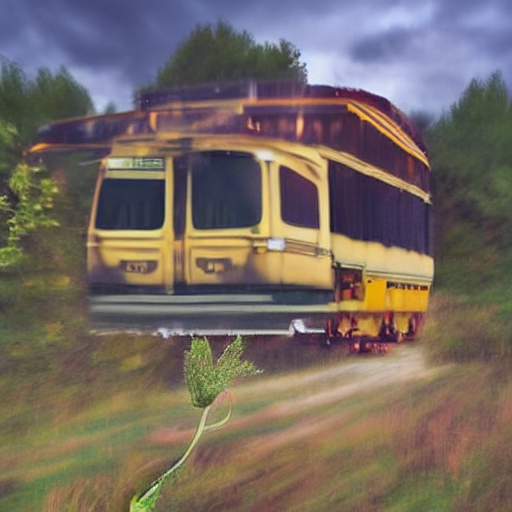}&\myim{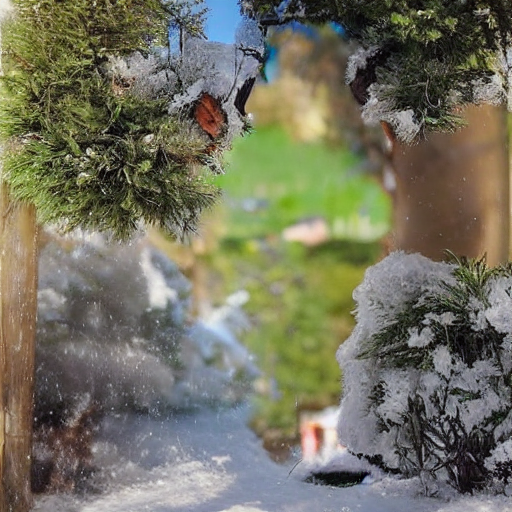} \\
\begin{sideways} \hspace{0.9cm} PwW\end{sideways} &
\myim{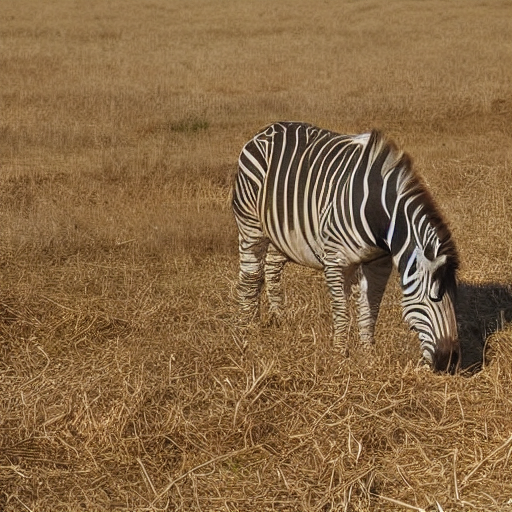}& 
\myim{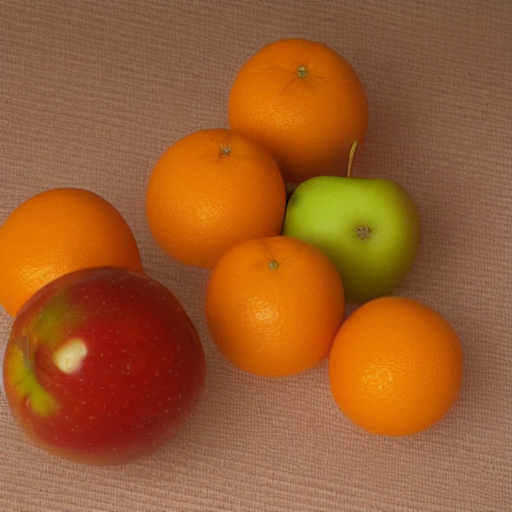}& 
\myim{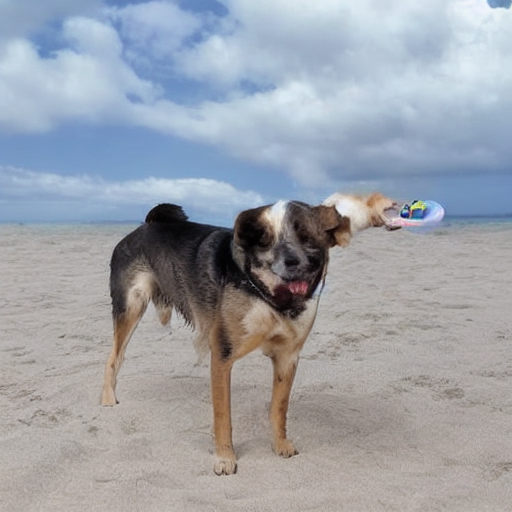}& \myim{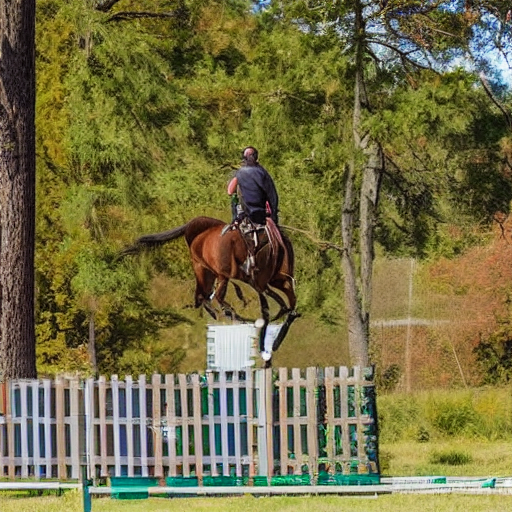}& \myim{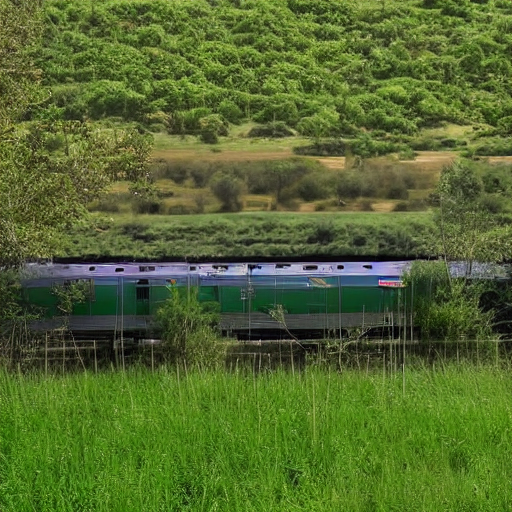}& \myim{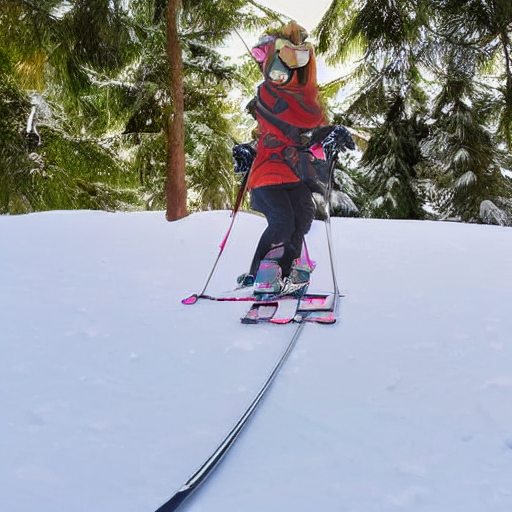}\\
\begin{sideways}\hspace{0.2cm}\ours (ours) \end{sideways} &
\myim{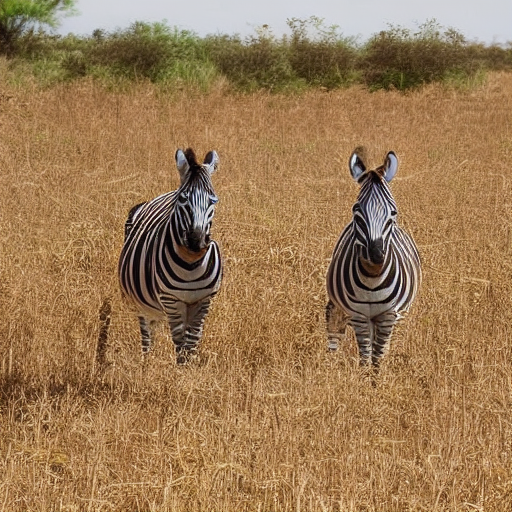}& 
\myim{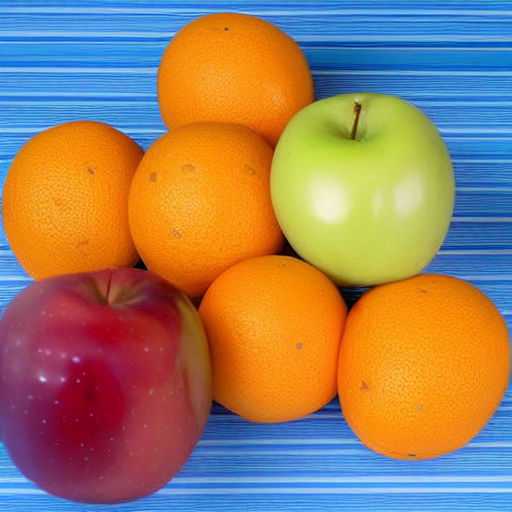}& 
\myim{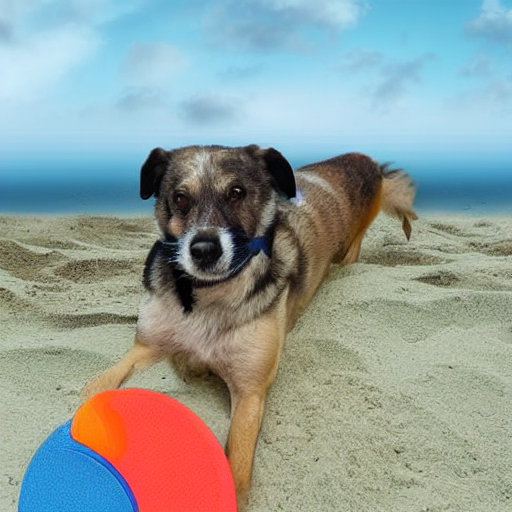}& \myim{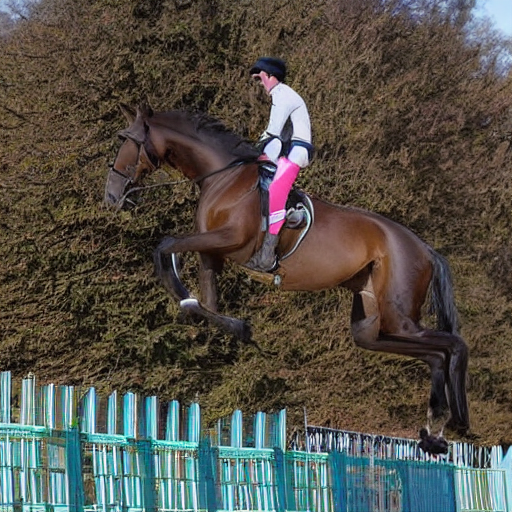}& \myim{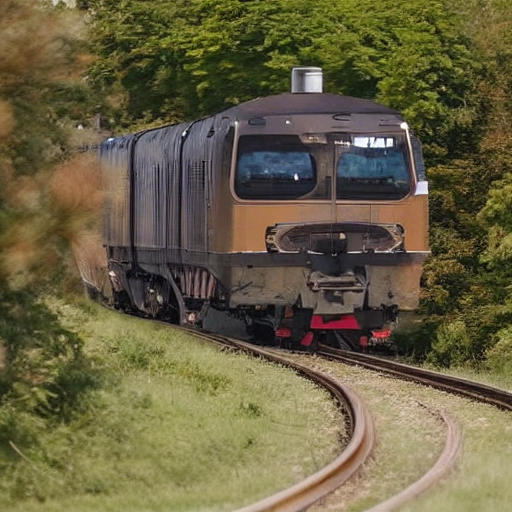}& \myim{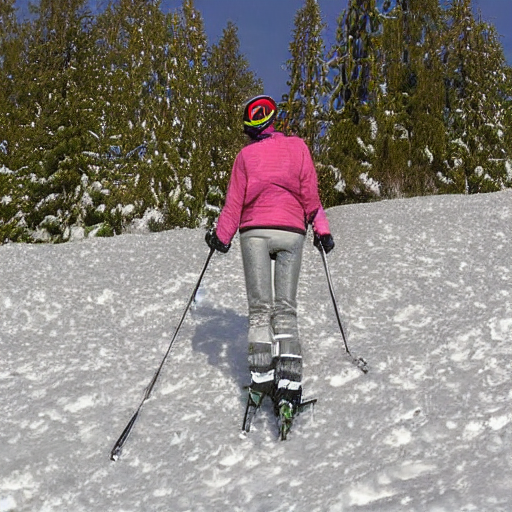}
\end{tabular}
\myvspace
\caption{Qualitative comparison of \ours to other methods based on LDM, conditioning on COCO  captions and up to three segments. 
}
\label{fig:qualitative}
\end{figure*}

\subsection{Main results}
\label{sec:results}

We present our  evaluation results in \tab{evalAllCls}.
Compared  to other methods that allow free-text annotation of segments (bottom part of the table), our approach leads to marked improvements in mIoU in all settings. For example improving by more than 10 points (36.3 to 46.9) over the closest competitor PwW, in the most realistic Eval-few setting.
Note that we even improve over SpaText, which finetunes Stable Diffusion specifically for this task.  
In terms of  CLIP score,  our approach yields similar or better results across all settings.  
Our approach obtains the best FID values among the methods based on our LDM text-to-image model. 
SpaText obtains the best overall FID values, which we attribute to the fact that it is finetuned on a dataset very similar to COCO, unlike the vanilla Stable Diffusion or our LDM.

In the top part of the table we report  results for methods that do not allow to condition segments on free-form text, and all require training on images with semantic segmentation maps. 
We find they perform well in the  Eval-all setting for which they are trained, and also in the similar  Eval-filtered setting, but deteriorate in the  Eval-few setting where only a few segments are provided as input.
In the Eval-few setting, our \ours approach  surpasses all  methods in the top part of the table in terms of mIoU.
Compared to LDM w/ External Classfier, which is based on the same diffusion model as \ours but does not allow to condition segments on free text, we improve across all metrics and settings, while being much faster at inference: 
LDM w/ ExternalClassifier takes 1 min.\ for one image while \ours takes around 15 secs. 

We provide qualitative results for the methods based on LDM in \fig{qualitative} when conditioning on up to three segments, corresponding to the Eval-few setting.
Our \ours clearly leads to superior aligment between the conditioning masks and the generated content.

\subsection{Ablations}
\label{sec:ablations}

In this section we focus on evaluation settings \textit{Eval-filtered} and \textit{Eval-few}, which better reflect practical use cases. 
To reduce compute, metrics are computed with a subset of 2k images from the COCO val set.

\begin{figure}
    \centering
    \includegraphics[width=\linewidth]{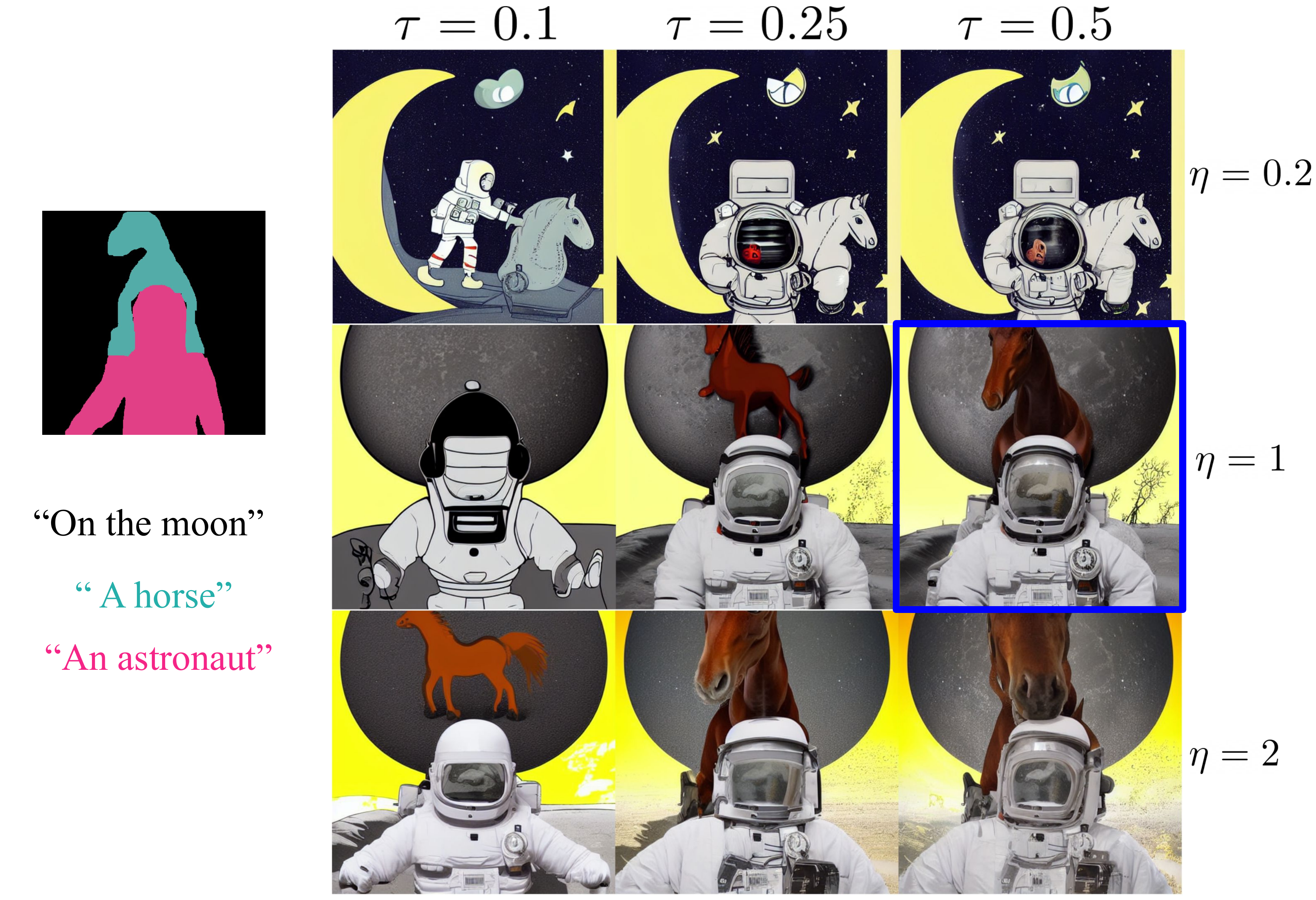}
    \caption{\ours outputs when varying the two main hyperparameters $\eta$ (learning rate) and $\tau$ (percentage of steps using classifier guidance). 
    Our default configuration is $\eta\!=\!1,\tau\!=\!0.5$.}
    \label{fig:viz_abl}
\end{figure}

\mypar{Ablation on hyperparameters $\tau$ and $\eta$} 
Our approach has two hyperparamters that control the strength of the spatial guidance:  the learning rate $\eta$ and the percentage of denoising steps $\tau$ until which classifier guidance is applied. 
Varying these hyperparameters strikes different trade-offs between mIoU (better with stronger guidance) and FID (better with less guidance and thus less perturbation of the diffusion model). 
In \fig{viz_abl} we show generations for a few values of these parameters. 
We can see that, given the right learning rate,  applying gradient updates for as few as the first $25\%$ denoising steps can suffice to enforce the layout conditioning. 
This is confirmed by quantitative results in the Eval-few setting presented in the supplementary material.
For $\eta\!=\!1$, setting $\tau\!=\!0.5$ strikes a good trade-off with an mIoU of $43.3$ and FID of $31.5$.  
Setting $\tau=1$ marginally improves mIoU by 1.3 points,  while worsening FID by 3.2 points, while setting $\tau=0.1$ worsens mIoU by 9.1 points for a gain of 1 point in FID. 
Setting $\tau\!=\!0.5$ requires additional compute for just the first half of denoising steps, making our method in practice only roughly 50\% more expensive than regular DDIM sampling.

\begin{figure}
    \centering
    \includegraphics[width=\linewidth]{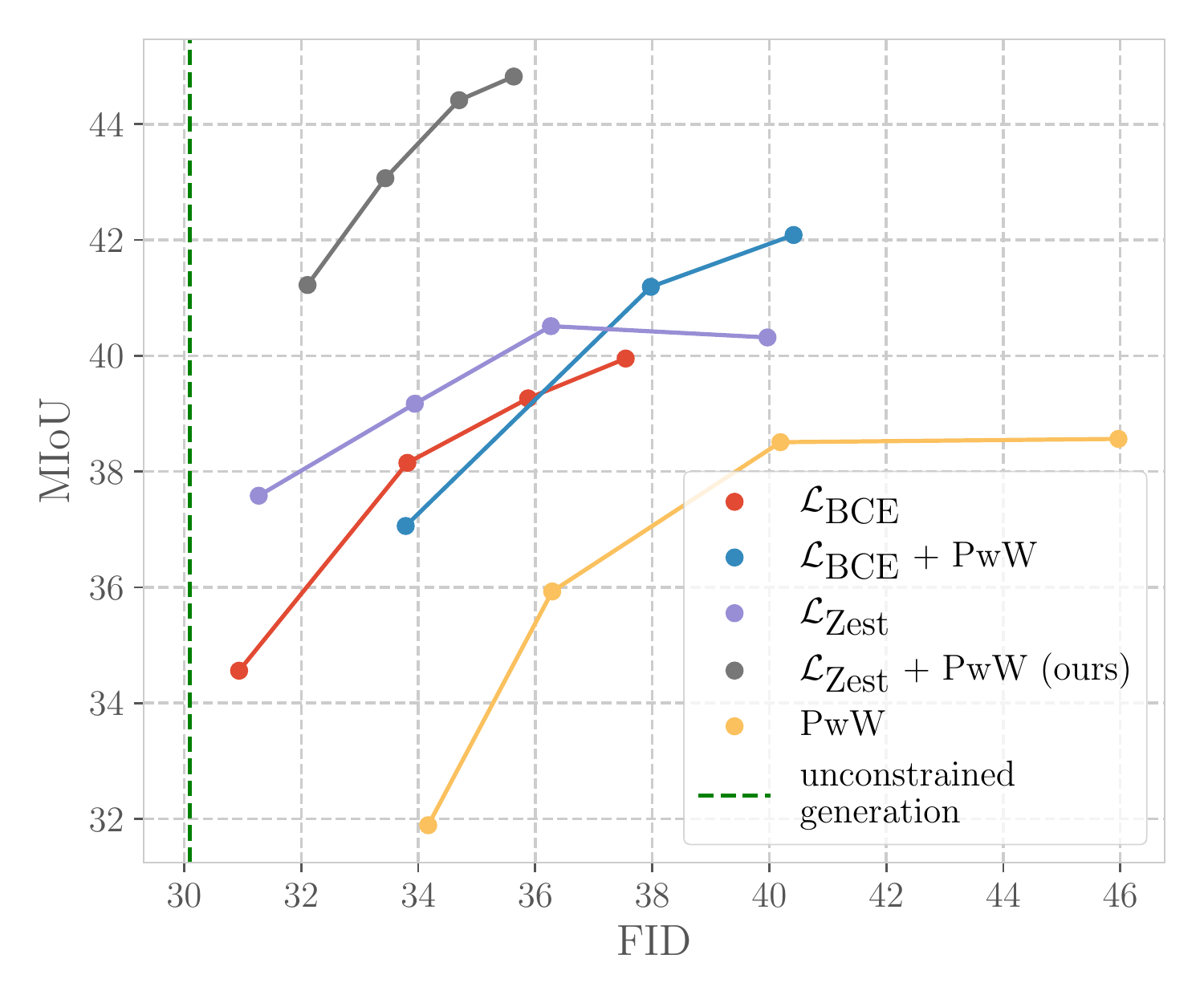}
    \caption{Trade-off in \textit{Eval-filtered} setting between FID (lower is better) and mIoU (higher is better) of PwW and \ours using different losses. In dotted green is shown the FID for unconstrained text-to-image generation. Using $\mathcal{L}_\textrm{Zest}$ in combination with PwW (our default setting) gives the best trade-off.
    }
    \label{fig:miou_fid}
\end{figure}

\mypar{Guidance losses and synergy with PwW} 
In \fig{miou_fid} we explore the  FID-mIoU trade-off in the Eval-filtered setting, for PwW and variations of our approach using different losses and with/out including PwW.
 The combined loss refers to our full loss in \Eq{segmentationloss}, while the BCE loss ignores the second normalized loss. 
For PwW, the FID-mIoU trade-off is controlled by the constant  $W$ that is added to the attention values to reinforce the association of image regions and their corresponding text.
For \ours, we vary $\eta$ to obtain different trade-offs, with $\tau\!=\!0.5$.
We observe that all versions of our approach provide  better mIoU-FID trade-offs than PwW alone. 
Interestingly, using the combined loss and PwW separately hardly improve the mIoU-FID trade-off \wrt only using the BCE loss, but their combination  gives a much better trade-off (Combined Loss + pWW).  
This is possibly due  to the  loss with normalized maps helping to produce more uniform segmentation masks, which helps PwW to provide more consistent updates.

In the remainder of the ablations, we consider the simplest version of \ours with the $\mathcal{L}_\mathrm{BCE}$ loss and without PwW, to better isolate the effect of gradient guiding.

\mypar{Attention map averaging} 
As mentioned in \sect{zeroshot}, we found that averaging the attention maps  across all heads of the different cross-attention layers is important to obtain good spatial localization. 
We review this choice in \tab{averaging}. 
When we compute our loss on each head separately, we can see a big drop in mIoU scores (-11 points). 
This reflects our observation that each attention head focuses on different parts of each object. 
By computing a loss on the averaged maps, a global pattern is enforced while still maintaining flexibility for each attention head.
This effect is much less visible when we average attention maps per layer, and apply the loss per layer: in this case mIoU deteriorates by 1.6 points, while FID improves by 0.9 points. %

\begin{table}
\centering
 \setlength{\tabcolsep}{5pt} 
 {\small
\begin{tabular}{lccc}
\toprule
\textbf{Components} &    $\downarrow$\textbf{FID}  &    $\uparrow$\textbf{mIoU} &    $\uparrow$\textbf{CLIP}  \\  
\midrule
Loss for each attention head   & 33.6 & 32.1 & 29.9 \\
Loss for each layer            & 31.6 & 42.7 & 30.5 \\
Loss for global average (ours) & 31.5 & 43.3 & 30.4 \\
\bottomrule
\end{tabular}
}
\myvspace
\caption{Evaluation of \ours on Eval-few setting, with different averaging schemes for computing the loss. 
Averaging all attention heads before applying the loss gives best results.}
\label{tab:averaging}
\end{table}

\mypar{Gradient normalization} Unlike standard classifier guidance, \ours uses normalized gradient to harmonize gradient descent updates in \Eq{ZestUpdate}. 
We find that while \ours also works without normalizing gradient, adding it gives a boost of 2 mIoU points for comparable FID scores. Qualitatively, it helped for some cases where the gradient norm was too high at the beginning of generation process, which occasionally resulted in low-quality samples.

Additional ablations are provided in the supplementary. 

\section{Conclusion}

In this paper, we have presented \ours, a zero-shot method which enables precise spatial control over the generated content by conditioning on segmentation masks annotated with free-form textual descriptions.
 Our approach leverages implicit segmentation maps extracted from text-attention in pre-trained text-to-image diffusion models to align the generation with input masks. Experimental results demonstrate that our approach achieves high-quality image generation while accurately aligning the generated content with input segmentations. Our quantitative evaluation shows that \ours is even competitive with methods trained on large image-segmentation datasets. Despite this success, there remains a limitation shared by many existing approaches. Specifically, the current approach, like others, tends to overlook small objects in the input conditioning maps. Further  work is required to address this problem which may be related to the low resolution of the attention maps in the diffusion model. 

\ificcvfinal
\paragraph{Acknowledgments.}
We would like to thank Oron Ashual, Uriel Singer, Adam Polyak and Shelly Sheynin for preparing the data and training and sharing the text-to-image model on which the work in this paper is based. 
\fi

{\small
\bibliographystyle{ieee_fullname}
\bibliography{egbib,jjv}}

\begin{thebibliography}{10}\itemsep=-1pt

\bibitem{avrahami2022spatext}
Omri Avrahami, Thomas Hayes, Oran Gafni, Sonal Gupta, Yaniv Taigman, Devi
  Parikh, Dani Lischinski, Ohad Fried, and Xi Yin.
\newblock {SpaText}: Spatio-textual representation for controllable image
  generation.
\newblock {\em arXiv preprint}, arXiv:2211.14305, 2022.

\bibitem{balaji22}
Yogesh Balaji, Seungjun Nah, Xun Huang, Arash Vahdat, Jiaming Song, Karsten
  Kreis, Miika Aittala, Timo Aila, Samuli Laine, Bryan Catanzaro, Tero Karras,
  and Ming-Yu Liu.
\newblock {eDiff-I}: Text-to-image diffusion models with ensemble of expert
  denoisers.
\newblock {\em arXiv preprint}, arXiv:2211.01324, 2022.

\bibitem{bansal2023universal}
Arpit Bansal, Hong-Min Chu, Avi Schwarzschild, Soumyadip Sengupta, Micah
  Goldblum, Jonas Geiping, and Tom Goldstein.
\newblock Universal guidance for diffusion models.
\newblock {\em arXiv preprint}, arXiv:2302.07121, 2023.

\bibitem{bartal23}
Omer Bar-Tal, Lior Yariv, Yaron Lipman, and Tali Dekel.
\newblock {MultiDiffusion}: Fusing diffusion paths for controlled image
  generation.
\newblock {\em arXiv preprint}, arXiv:2302.08113, 2023.

\bibitem{caesar18cvpr}
Holger Caesar, Jasper Uijlings, and Vittorio Ferrari.
\newblock {COCO-Stuff}: Thing and stuff classes in context.
\newblock In {\em CVPR}, 2018.

\bibitem{chang22cvpr}
Huiwen Chang, Han Zhang, Lu Jiang, Ce Liu, and William~T. Freeman.
\newblock {MaskGIT}: Masked generative image transformer.
\newblock In {\em CVPR}, 2022.

\bibitem{chefer2023attend}
Hila Chefer, Yuval Alaluf, Yael Vinker, Lior Wolf, and Daniel Cohen-Or.
\newblock Attend-and-excite: Attention-based semantic guidance for
  text-to-image diffusion models.
\newblock {\em arXiv preprint}, arXiv:2301.13826, 2023.

\bibitem{chen15iclr}
L.-C. Chen, G. Papandreou, I. Kokkinos, K. Murphy, and A. Yuille.
\newblock Semantic image segmentation with deep convolutional nets and fully
  connected {CRF}s.
\newblock In {\em ICLR}, 2015.

\bibitem{chen2017deeplab}
Liang-Chieh Chen, George Papandreou, Iasonas Kokkinos, Kevin Murphy, and Alan~L
  Yuille.
\newblock {DeepLab}: Semantic image segmentation with deep convolutional nets,
  atrous convolution, and fully connected {CRF}s.
\newblock {\em PAMI}, 40(4):834--848, 2017.

\bibitem{chen2022vitadapter}
Zhe Chen, Yuchen Duan, Wenhai Wang, Junjun He, Tong Lu, Jifeng Dai, and Yu
  Qiao.
\newblock Vision transformer adapter for dense predictions.
\newblock In {\em ICLR}, 2023.

\bibitem{couairon22}
Guillaume Couairon, Jakob Verbeek, Holger Schwenk, and Matthieu Cord.
\newblock {DiffEdit}: Diffusion-based semantic image editing with mask
  generation.
\newblock In {\em ICLR}, 2023.

\bibitem{dhariwal21nips}
Prafulla Dhariwal and Alex Nichol.
\newblock Diffusion models beat {GAN}s on image synthesis.
\newblock In {\em NeurIPS}, 2021.

\bibitem{esser21cvpr}
Patrick Esser, Robin Rombach, and B. Ommer.
\newblock Taming transformers for high-resolution image synthesis.
\newblock In {\em CVPR}, 2021.

\bibitem{feng2022training}
Weixi Feng, Xuehai He, Tsu-Jui Fu, Varun Jampani, Arjun Akula, Pradyumna
  Narayana, Sugato Basu, Xin~Eric Wang, and William~Yang Wang.
\newblock Training-free structured diffusion guidance for compositional
  text-to-image synthesis.
\newblock {\em arXiv preprint}, arXiv:2212.05032, 2022.

\bibitem{gafni22arxiv}
Oran Gafni, Adam Polyak, Oron Ashual, Shelly Sheynin, Devi Parikh, and Yaniv
  Taigman.
\newblock Make-a-scene: Scene-based text-to-image generation with human priors.
\newblock In {\em ECCV}, 2022.

\bibitem{hertz2022prompt}
Amir Hertz, Ron Mokady, Jay Tenenbaum, Kfir Aberman, Yael Pritch, and Daniel
  Cohen-Or.
\newblock Prompt-to-prompt image editing with cross attention control.
\newblock {\em arXiv preprint}, arXiv:2208.01626, 2022.

\bibitem{heusel17nips}
Martin Heusel, Hubert Ramsauer, Thomas Unterthiner, Bernhard Nessler, and Sepp
  Hochreiter.
\newblock {GAN}s trained by a two time-scale update rule converge to a local
  {N}ash equilibrium.
\newblock In {\em NeurIPS}, 2017.

\bibitem{ho20neurips}
Jonathan Ho, Ajay Jain, and Pieter Abbeel.
\newblock Denoising diffusion probabilistic models.
\newblock In {\em NeurIPS}, 2020.

\bibitem{isola17cvpr}
Phillip Isola, Jun-Yan Zhu, Tinghui Zhou, and Alexei~A. Efros.
\newblock Image-to-image translation with conditional adversarial networks.
\newblock In {\em CVPR}, 2017.

\bibitem{lezama22eccv}
Jos\'e Lezama, Huiwen Chang, Lu Jiang, and Irfan Essa.
\newblock Improved masked image generation with token-critic.
\newblock In {\em ECCV}, 2022.

\bibitem{li2023gligen}
Yuheng Li, Haotian Liu, Qingyang Wu, Fangzhou Mu, Jianwei Yang, Jianfeng Gao,
  Chunyuan Li, and Yong~Jae Lee.
\newblock {GLIGEN}: Open-set grounded text-to-image generation.
\newblock {\em arXiv preprint}, arXiv:2301.07093, 2023.

\bibitem{luddecke22cvpr}
Timo L\"uddecke and Alexander~S. Ecker.
\newblock Image segmentation using text and image prompts.
\newblock In {\em CVPR}, 2022.

\bibitem{ma2023directed}
Wan-Duo~Kurt Ma, JP Lewis, W~Bastiaan Kleijn, and Thomas Leung.
\newblock Directed diffusion: Direct control of object placement through
  attention guidance.
\newblock {\em arXiv preprint arXiv:2302.13153}, 2023.

\bibitem{meng2022sdedit}
Chenlin Meng, Yutong He, Yang Song, Jiaming Song, Jiajun Wu, Jun-Yan Zhu, and
  Stefano Ermon.
\newblock {SDE}dit: Guided image synthesis and editing with stochastic
  differential equations.
\newblock In {\em ICLR}, 2022.

\bibitem{mou2023t2i}
Chong Mou, Xintao Wang, Liangbin Xie, Jian Zhang, Zhongang Qi, Ying Shan, and
  Xiaohu Qie.
\newblock {T2I-Adapter}: Learning adapters to dig out more controllable ability
  for text-to-image diffusion models.
\newblock {\em arXiv preprint}, arXiv:2302.08453, 2023.

\bibitem{nichol21arxiv}
Alex Nichol, Prafulla Dhariwal, Aditya Ramesh, Pranav Shyam, Pamela Mishkin,
  Bob {McGrew}, Ilya Sutskever, and Mark Chen.
\newblock {GLIDE}: Towards photorealistic image generation and editing with
  text-guided diffusion models.
\newblock In {\em ICML}, 2022.

\bibitem{oord17nips}
A.~van~den Oord, O. Vinyals, and K. Kavukcuoglu.
\newblock Neural discrete representation learning.
\newblock In {\em NeurIPS}, 2017.

\bibitem{casanova22}
Arantxa~Casanova Paga, Marlene Careil, Adriana~Romero Soriano, Christopher~J.
  Pal, Jakob Verbeek, and Michal Drozdzal.
\newblock Controllable image generation via collage representations.
\newblock {\em ICLR submission}, 2022.

\bibitem{park19cvpr1}
T. Park, M.-Y. Liu, T.-C. Wang, and J.-Y. Zhu.
\newblock Semantic image synthesis with spatially-adaptive normalization.
\newblock In {\em CVPR}, 2019.

\bibitem{parmar2023zero}
Gaurav Parmar, Krishna~Kumar Singh, Richard Zhang, Yijun Li, Jingwan Lu, and
  Jun-Yan Zhu.
\newblock Zero-shot image-to-image translation.
\newblock {\em arXiv preprint}, arXiv:2302.03027, 2023.

\bibitem{parmar2021cleanfid}
Gaurav Parmar, Richard Zhang, and Jun-Yan Zhu.
\newblock On aliased resizing and surprising subtleties in {GAN} evaluation.
\newblock In {\em CVPR}, 2022.

\bibitem{radford21clip}
Alec Radford, Jong~Wook Kim, Chris Hallacy, Aditya Ramesh, Gabriel Goh,
  Sandhini Agarwal, Girish Sastry, Amanda Askell, Pamela Mishkin, Jack Clark,
  Gretchen Krueger, and Ilya Sutskever.
\newblock Learning transferable visual models from natural language
  supervision.
\newblock In {\em ICML}, 2021.

\bibitem{raffel20jmlr}
Colin Raffel, Noam Shazeer, Adam Roberts, Katherine Lee, Sharan Narang, Michael
  Matena, Yanqi Zhou, Wei Li, and Peter~J. Liu.
\newblock Exploring the limits of transfer learning with a unified text-to-text
  transformer.
\newblock {\em JMLR}, 21, 2022.

\bibitem{ramesh22dalle2}
A. Ramesh, P. Dhariwal, A. Nichol, C. Chu, and M. Chen.
\newblock Hierarchical text-conditionalimage generation with {CLIP} latents.
\newblock {\em arXiv preprint}, arXiv:2204.06125, 2022.

\bibitem{razavi19nips}
Ali Razavi, Aaron van~den Oord, and Oriol Vinyals.
\newblock Generating diverse high-fidelity images with {VQ-VAE-2}.
\newblock In {\em NeurIPS}, 2019.

\bibitem{rombach21arxiv}
Robin Rombach, Andreas Blattmann, Dominik Lorenz, Patrick Esser, and Bj\"orn
  Ommer.
\newblock High-resolution image synthesis with latent diffusion models.
\newblock In {\em CVPR}, 2022.

\bibitem{saharia2022palette}
Chitwan Saharia, William Chan, Huiwen Chang, Chris Lee, Jonathan Ho, Tim
  Salimans, David Fleet, and Mohammad Norouzi.
\newblock Palette: Image-to-image diffusion models.
\newblock In {\em ACM SIGGRAPH}, 2022.

\bibitem{saharia22nips}
Chitwan Saharia, William Chan, Saurabh Saxena, Lala Li, Jay Whang, Emily
  Denton, Seyed Kamyar~Seyed Ghasemipour, Burcu~Karagol Ayan, S.~Sara Mahdavi,
  Rapha~Gontijo Lopes, Tim Salimans, Jonathan Ho, David~J Fleet, and Mohammad
  Norouzi.
\newblock Photorealistic text-to-image diffusion models with deep language
  understanding.
\newblock In {\em NeurIPS}, 2022.

\bibitem{sushko21iclr}
Edgar Sch\"onfeld, Vadim Sushko, Dan Zhang, Juergen Gall, Bernt Schiele, and
  Anna Khoreva.
\newblock You only need adversarial supervision for semantic image synthesis.
\newblock In {\em ICLR}, 2021.

\bibitem{sohl-dickstein15icml}
Jascha Sohl-Dickstein, Eric Weiss, Niru Maheswaranathan, and Surya Ganguli.
\newblock Deep unsupervised learning using nonequilibrium thermodynamics.
\newblock In {\em ICML}, 2015.

\bibitem{song2020denoising}
Jiaming Song, Chenlin Meng, and Stefano Ermon.
\newblock Denoising diffusion implicit models.
\newblock In {\em ICLR}, 2020.

\bibitem{song21iclr2}
Yang Song, Jascha Sohl-Dickstein, Diederik~P. Kingma, Abhishek Kumar, Stefano
  Ermon, and Ben Poole.
\newblock Score-based generative modeling through stochastic differential
  equations.
\newblock In {\em ICLR}, 2021.

\bibitem{vaswani17nips}
A. Vaswani, N. Shazeer, N. Parmar, J. Uszkoreit, L. Jones, A. Gomez, L. Kaiser,
  and I. Polosukhin.
\newblock Attention is all you need.
\newblock In {\em NeurIPS}, 2017.

\bibitem{wang22arxiv}
Tengfei Wang, Ting Zhang, Bo Zhang, Hao Ouyang, Dong Chen, Qifeng Chen, and
  Fang Wen.
\newblock Pretraining is all you need for image-to-image translation.
\newblock {\em arXiv preprint}, arXiv:2205.12952, 2022.

\bibitem{wang2022semantic}
Weilun Wang, Jianmin Bao, Wengang Zhou, Dongdong Chen, Dong Chen, Lu Yuan, and
  Houqiang Li.
\newblock Semantic image synthesis via diffusion models.
\newblock {\em arXiv preprint}, arXiv:2207.00050, 2022.

\bibitem{zhang2023adding}
Lvmin Zhang and Maneesh Agrawala.
\newblock Adding conditional control to text-to-image diffusion models.
\newblock {\em arXiv preprint}, arXiv:2302.05543, 2023.

\end{thebibliography}

\clearpage\appendix

\section{Societal impact}
Our work advances the capabilities of generative image models, contributing to the democratization of creative design by offering tools for non-expert users.
Generative image models, however, also pose risks, including  using these tools to generate harmful content or deep-fakes, or models generating images similar to the training data which may contain personal data. 
These concerns have led us to steer away from using large-scale open-source generative image models trained on datasets scraped from the web, for which the licensing of the content is not always clear and which may contain harmful content.  Instead, we  trained models on a large in-house curated dataset which mitigates  these concerns to a large extent.

\section{Implementation details }

\mypar{Implementation details} 
For all experiments that use our LDM diffusion model, we use 50 steps of DDIM sampling with classifier-free guidance strength set to 3. Stable Diffusion-based competing methods, like PwW, also use 50 steps of DDIM sampling, but with a classifier-free guidance of 7.5.

\mypar{Computation of metrics}
To compute the mIoU metric we use ViT-Adapter\cite{chen2022vitadapter} as segmentation model rather than the commonly used DeepLabV2~\cite{chen15iclr}, as the former improves over the latter by 18.6  points of mIoU (from 35.6 to 54.2) on COCO-Stuff. Scores for methods based on Stable Diffusion are taken from \url{https://cdancette.fr/diffusion-models/}.

\section{Additional ablation experiments}

For these additional ablation experiments, we use the \textit{Eval-few} setting as presented in the paper, where $1\leq K \leq 3$ spatial masks are used for conditioning.

\mypar{Attention layers used} 
We first validate which layers are useful for computing our classifier guidance loss in Table~\ref{abl_which_layer}. 
We  find that whatever the set of cross-attention layers used for computing loss, the mIoU and FID scores are very competitive. In accordance with preliminary observations, it is slightly better to skip attention maps at resolution 8 when computing our loss.

\begin{table}[H]
\centering
 \setlength{\tabcolsep}{5pt} 
{\small
\begin{tabular}{lccc}
\toprule
\textbf{Layers used} &    $\downarrow$\textbf{FID}  &    $\uparrow$\textbf{mIoU} &    $\uparrow$\textbf{CLIP}  \\  
\midrule
     All layers & 33.74 & 40.17 &      30.19 \\
 Only decoder layers & 33.81 & 40.02 &      30.05 \\
Only encoder layers & 30.98 & 38.24 &      30.67 \\
  Only res32 layers & \bf 29.35 & 39.49 &     \bf 30.75 \\
  Only res16 layers & 33.59 & 40.27 &      30.23 \\
      res16 and res32 layers (ours) & 31.53 & \bf 43.34 &      30.44 \\
\bottomrule
\end{tabular}
}
\caption{Ablation on cross-attention layers used for estimating segmentation maps.}
\label{abl_which_layer}
\end{table}

\mypar{Gradient normalization} We validate the impact of normalizing gradient when applying classifier guidance with our $\mathcal{L}_\textrm{Zest}$ loss. Results are in Table \ref{tab:gradnorm}.

\begin{table}
\centering
 \setlength{\tabcolsep}{5pt} 
{\small
\begin{tabular}{lccc}
\toprule
\textbf{Normalization} &    $\downarrow$\textbf{FID}  &    $\uparrow$\textbf{mIoU} &    $\uparrow$\textbf{CLIP}  \\  
\midrule
No normalization & 30.77 & 38.99 &      30.70 \\
L2 norm & \bf 28.57 & 36.39 &     \bf  31.27 \\
L1 norm & 28.85 & 39.74 &      31.04 \\
L$_\infty$ norm (ours) & 31.53 & \bf 43.34 &      30.44 \\
\bottomrule
\end{tabular}
}
\caption{Impact of gradient normalization scheme on performance.}
\label{tab:gradnorm}
\end{table}

\mypar{Impact of parameter $\tau$} In our method, classifier guidance is only used in a fraction $\tau$ of denoising steps, after which it is disabled. Table \ref{abl_tau} demonstrates that after our default value $\tau=0.5$, mIoU gains are marginal, while the FID scores are worse. Conversely, using only $10\%$ or $25\%$ of denoising steps for classifier guidance already gives very good mIoU/FID scores, better than PwW for $\tau=0.25$. As illustrated in \sect{add_viz}, this is because estimated segmentation maps converge very early in the generation process.

\begin{table}[H]
\centering
 \setlength{\tabcolsep}{5pt} 
{\small
\begin{tabular}{lccc}
\toprule
\textbf{Components} &    $\downarrow$\textbf{FID}  &    $\uparrow$\textbf{mIoU} &    $\uparrow$\textbf{CLIP}  \\  
\midrule
$\tau=0.1$& 30.54 & 34.25 &     \bf  31.18 \\
$\tau=0.25$ & \bf 30.36 & 40.75 &      30.77 \\
$\tau=0.5$ & 31.53 & 43.34 &      30.44 \\
  $\tau=1$ & 34.75 & \bf 44.58 &      29.99 \\
  
\bottomrule
\end{tabular}
}
\caption{Ablation on parameter $\tau$, with fixed learning rate $\eta=1$ in the Eval-few setting.}
\label{abl_tau}
\end{table}

\mypar{Tokens used as attention keys}
Our estimated segmentation masks are computed with an attention mechanism over a set of keys computed from  the text prompt embeddings. 
In this experiment, we analyze whether the attention over the full text-prompt is necessary, or whether we could simply use classification scores over the set of classes corresponding to the segments. We encode each class text separately with the text encoder, followed by average pooling to get a single embedding per class. Computing our loss with these embeddings as attention keys results in a probability distribution over the segmentation classes. We find that the FID scores are worse (+ 3 pts FID), but the mIoU scores are very close  (43.36 \vs 43.34). We conclude that our loss function primarily serves to align spatial image features with the relevant textual feature at each spatial location, and that the patterns that we observe in attention maps are  a manifestation of this alignment.

\begin{figure*}

\centering
\small
\setlength{\tabcolsep}{1pt}

\def\myim#1{\includegraphics[width=30mm,height=30mm]{figs/supmat/#1.png}}

\begin{tabular}{ccc} 
``A big burly grizzly \\
bear is shown with grass \\
in the background.'' \\
\myim{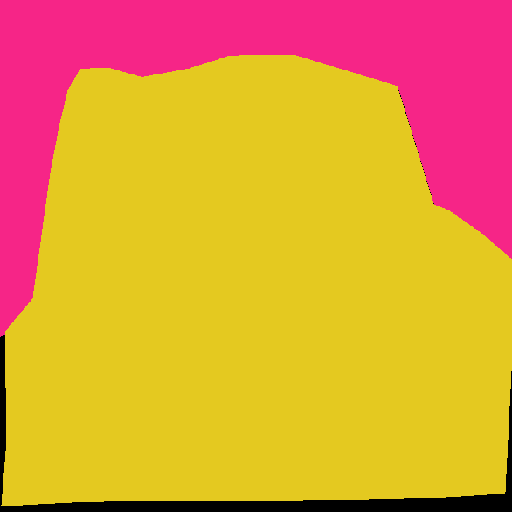}& \myim{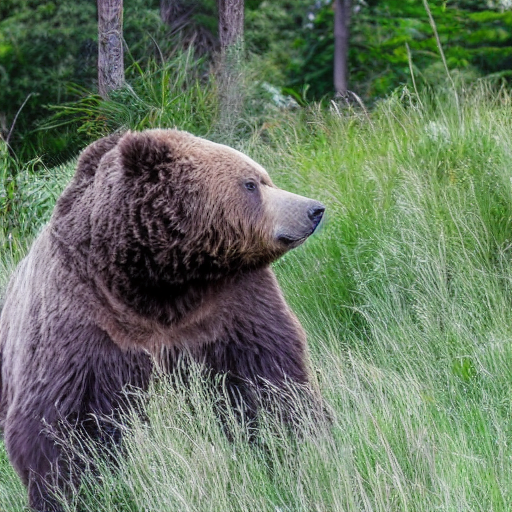} & \myim{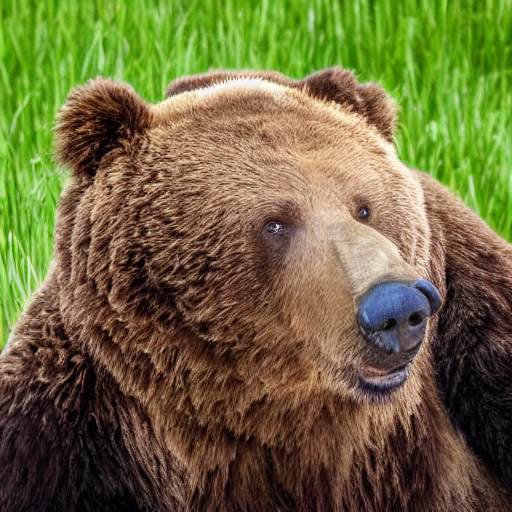}   \\
 \red{grass} &W/o Guidance & W/ Guidance \\
 \yel{bear} &\\
\end{tabular}

\def\myim#1{\includegraphics[width=17cm]{figs/supmat/#1.png}}

\begin{tabular}{cc} 
\multicolumn{2}{c}{W/o Guidance} \\
\begin{sideways} \hspace{0.5cm} Bear\end{sideways}& \myim{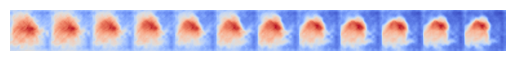} \\
\begin{sideways}  \hspace{0.5cm} Grass\end{sideways}& \myim{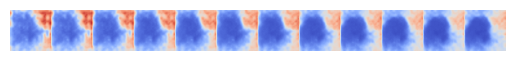} \\

\multicolumn{2}{c}{W/ Guidance} \\
\begin{sideways}\hspace{0.5cm} Bear\end{sideways}& \myim{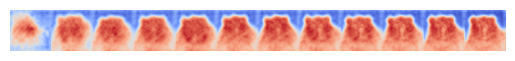} \\
\begin{sideways}\hspace{0.5cm}Grass\end{sideways}& \myim{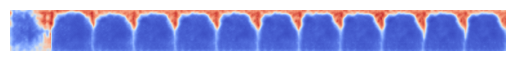} \\
\end{tabular}
\vspace{1mm}
\caption{
Visualization of first 12 denoising steps out of 50 steps. Same seed for w/ and w/o guidance. 
}
\label{fig:attention_guidance_bear}
\end{figure*}

\begin{figure*}

\centering
\small
\setlength{\tabcolsep}{1pt}

\def\myim#1{\includegraphics[width=30mm,height=30mm]{figs/supmat/#1.png}}
\def\myimMask#1{\includegraphics[width=30mm,height=30mm]{#1}}

\begin{tabular}{ccc} 
``Five oranges \\
with a red apple \\
and a green apple.'' \\
\myimMask{figs/freeform_cond/masks/534.png}& \myim{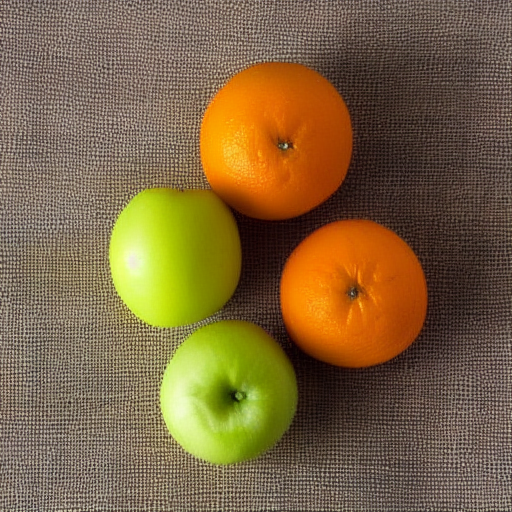} & \myim{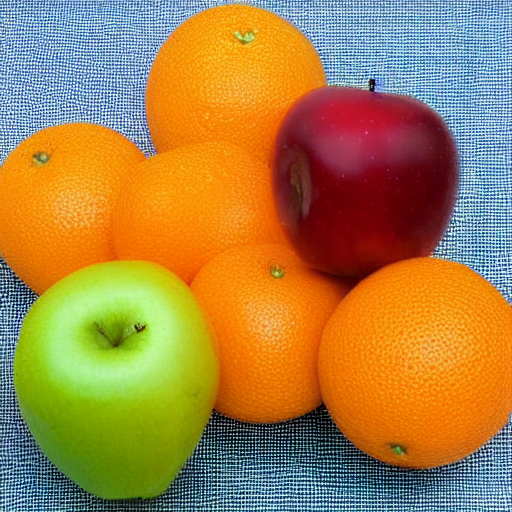}   \\
 \red{apple} &W/o Guidance & W/ Guidance \\
 \yel{orange} &\\
\end{tabular}

\def\myim#1{\includegraphics[width=17cm]{figs/supmat/#1.png}}

\begin{tabular}{cc} 
\multicolumn{2}{c}{W/o Guidance} \\
\begin{sideways} \hspace{0.5cm}Orange\end{sideways}& \myim{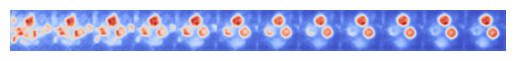} \\
\begin{sideways}  \hspace{0.5cm}Apple\end{sideways}& \myim{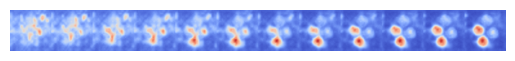} \\

\multicolumn{2}{c}{W/ Guidance} \\
\begin{sideways}\hspace{0.5cm} Orange\end{sideways}& \myim{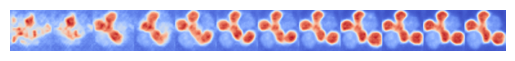} \\
\begin{sideways}\hspace{0.5cm}Apple\end{sideways}& \myim{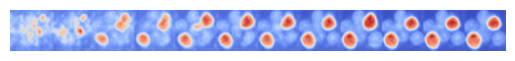} \\
\end{tabular}
\vspace{1mm}
\caption{
Visualization of first 12 denoising steps out of 50 steps. Same seed for w/ and w/o guidance. 
}
\label{fig:attention_guidance_fruit}
\end{figure*}

\begin{figure*}
 \def\myim#1{\includegraphics[width=27mm,height=27mm]{figs/supmat/coco_fig/#1}}
 \footnotesize
     \centering
   \setlength\tabcolsep{0.5 pt}
   \renewcommand{\arraystretch}{0.2}
     \begin{tabular}{ccccccc}

 &``6 open umbrellas    &``Cat sitting up   &  ``There are two brown   & ``A close-up of an   & ``A broken suitcase     & ``A group of zebras  \\
 &of various colors &with a fake tie  &  bears that are playing&  orange on the side &  is on the side & walking away\\
 &hanging on a line'' & around it's neck.'' &  together in the water.''  & of the road.'' & of the road.'' & from trees.''\\
 &\myim{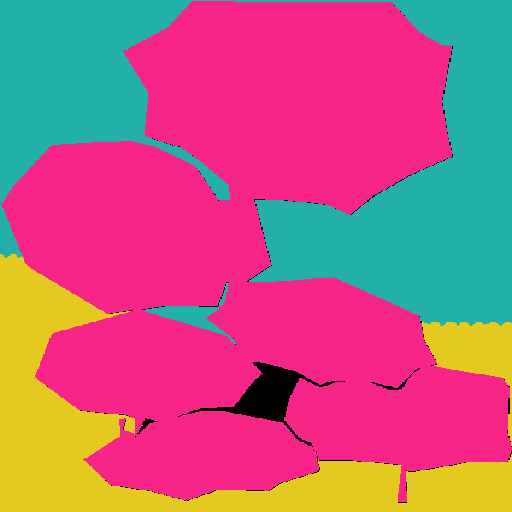} & 
\myim{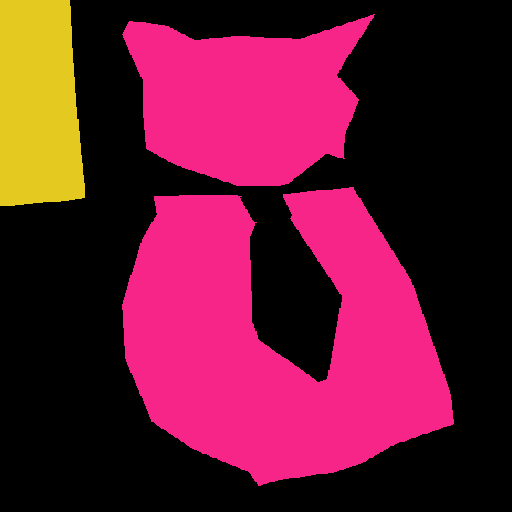}  & 
\myim{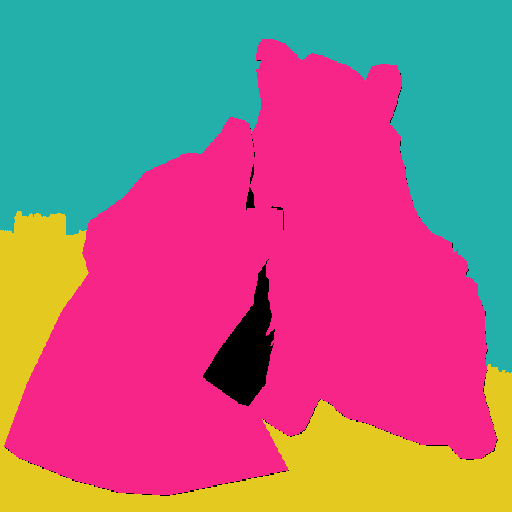}  &  \myim{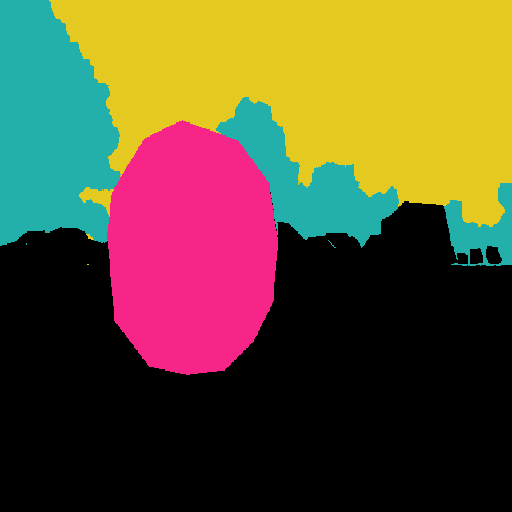} & \myim{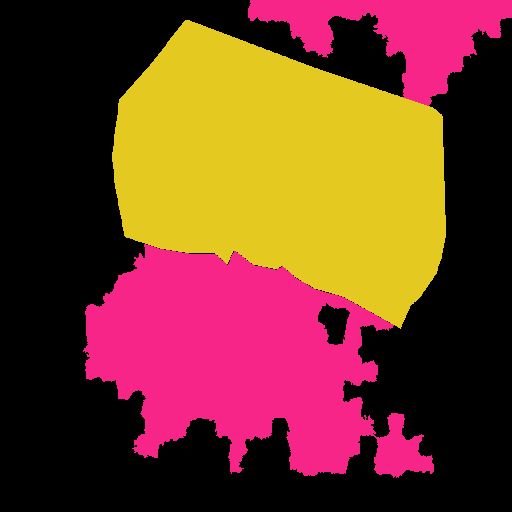}  & \myim{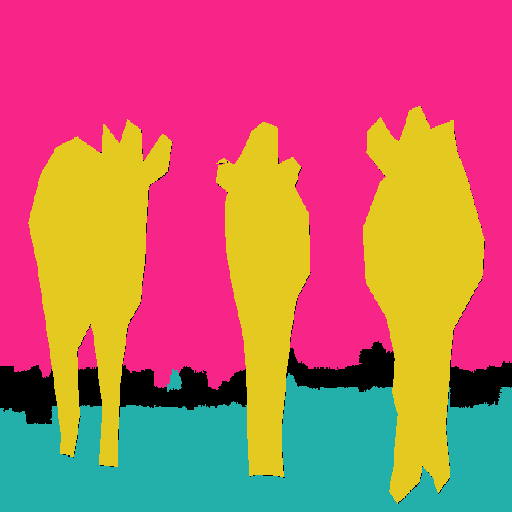}  \\

\vspace{0.5mm} \\
& \red{umbrella} & \red{cat} &\red{bear}  & \red{orange} & \red{plant} & \red{tree}\\
 &\yel{house}& \yel{furniture}& \yel{river}  & \yel{clouds} & \yel{suitcase} & \yel{zebra}\\

  &\blue{sky}& &\blue{wall}  &  \blue{tree} & \blue{wall} & \blue{dirt}\\
\begin{sideways}   \hspace{0.4cm}Ext. Classifier\end{sideways} &\myim{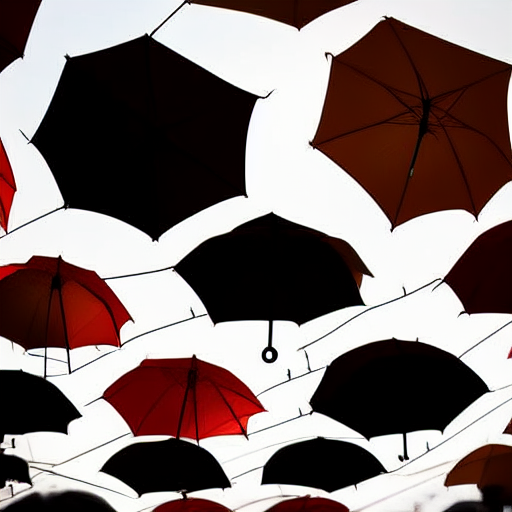}
&\myim{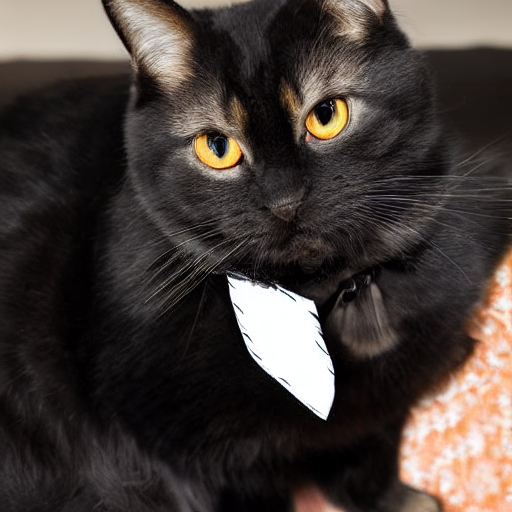}& 
\myim{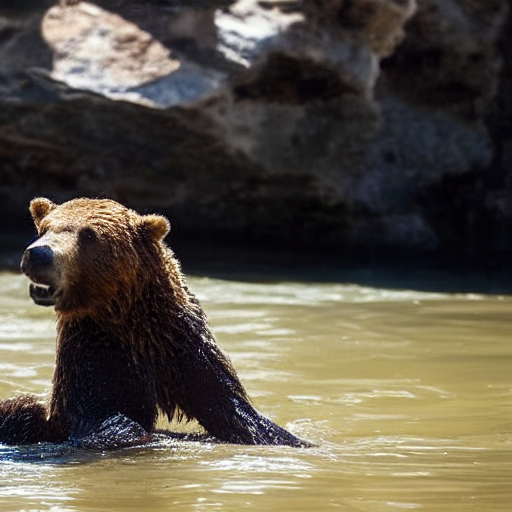}& \myim{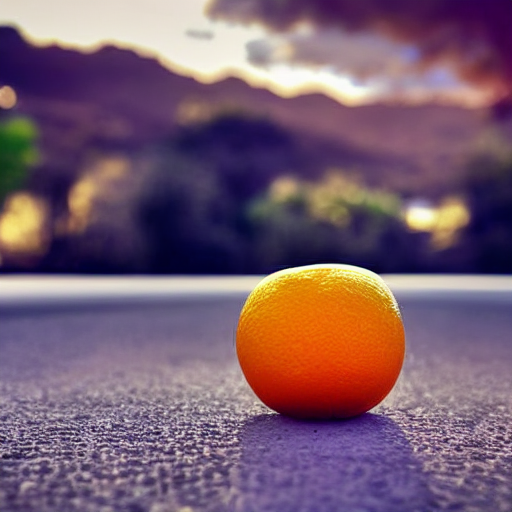}& \myim{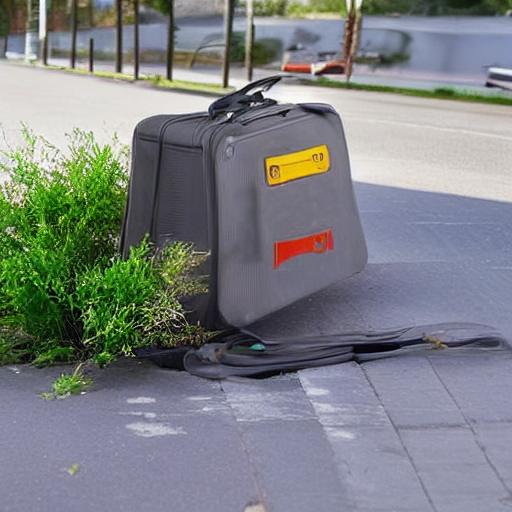}& \myim{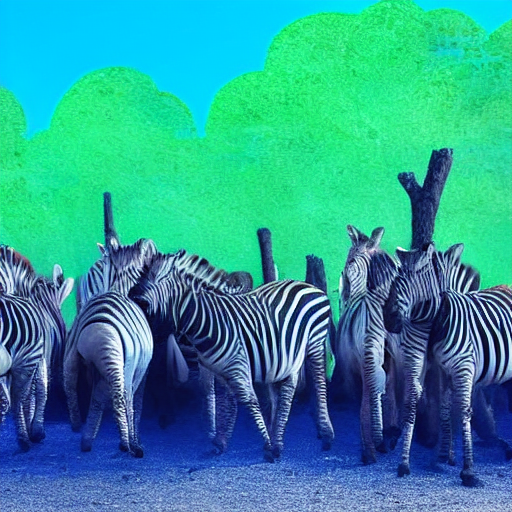} \\

\begin{sideways} \hspace{0.4cm}MultiDiffusion\end{sideways} &\myim{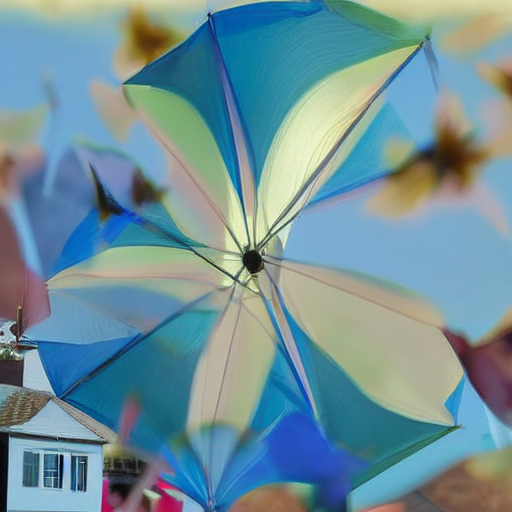}
&\myim{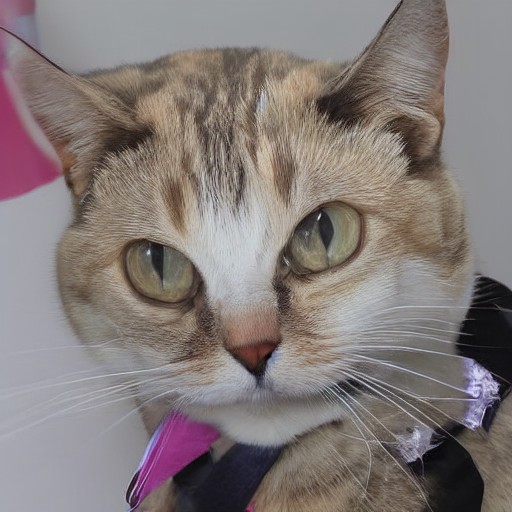}& 
\myim{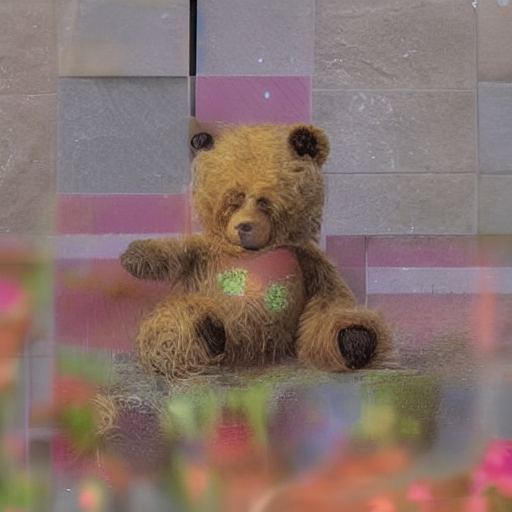}& \myim{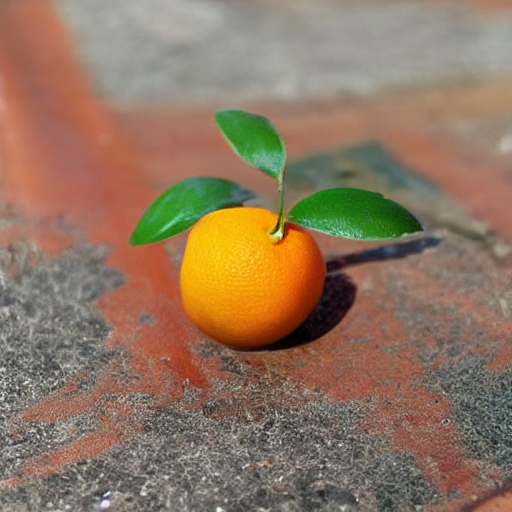}& \myim{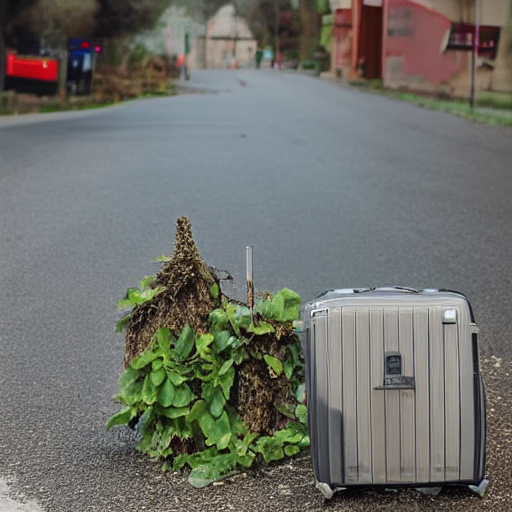}& \myim{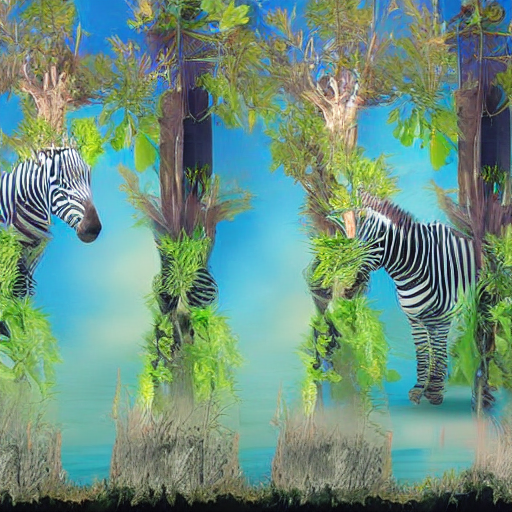} \\
\begin{sideways} \hspace{0.9cm} PwW\end{sideways} &\myim{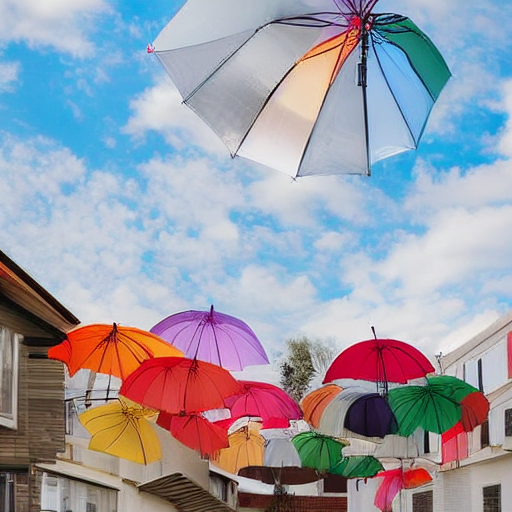}
&\myim{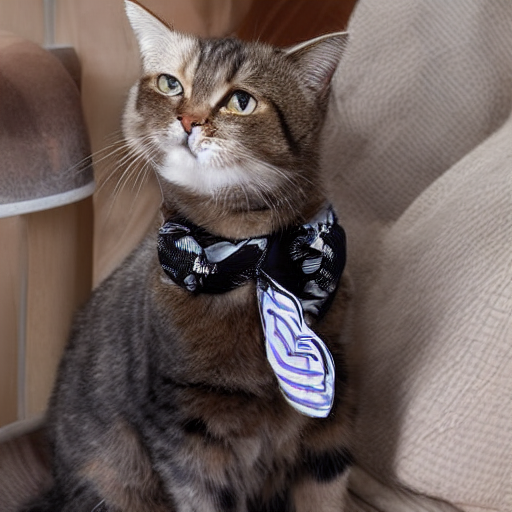}& 
\myim{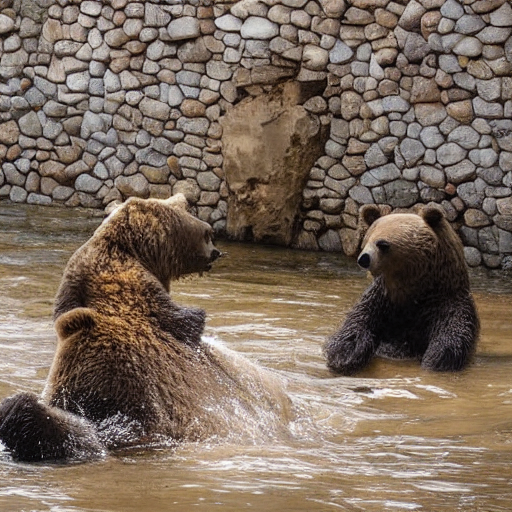}& \myim{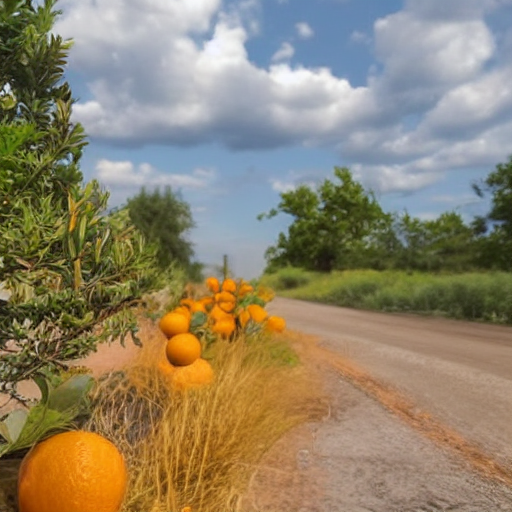}& \myim{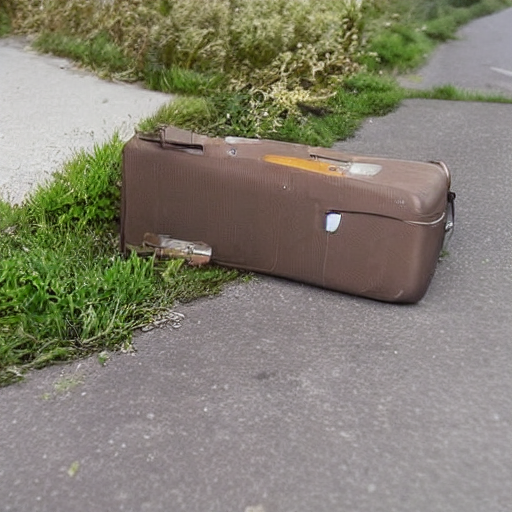}& \myim{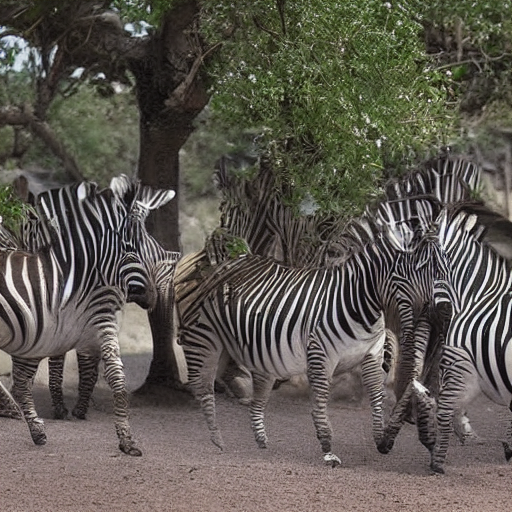} \\
\begin{sideways}\hspace{0.2cm}\ours (ours) \end{sideways} &\myim{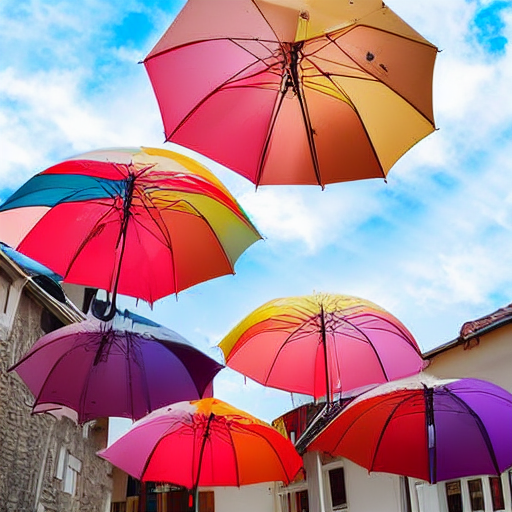}
&\myim{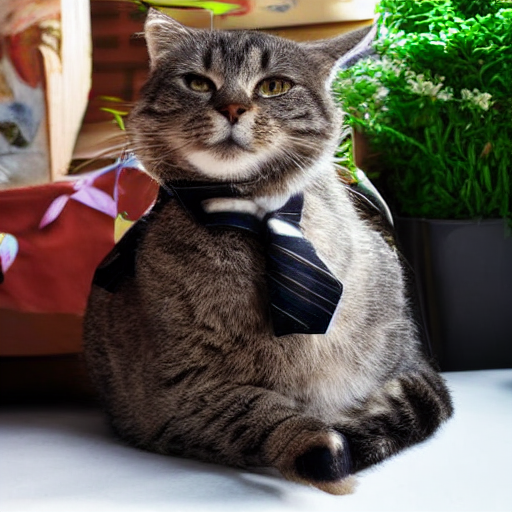}& 
\myim{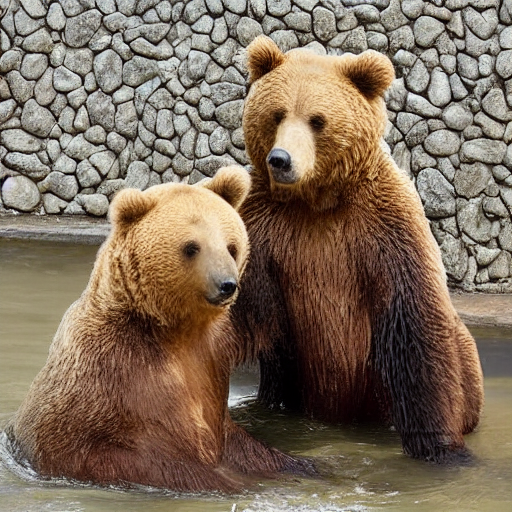}& \myim{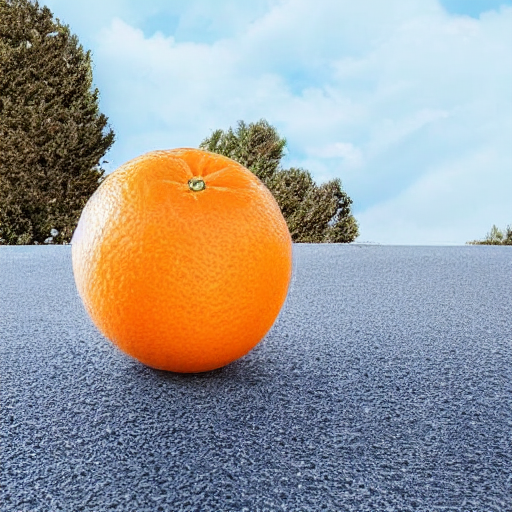}& \myim{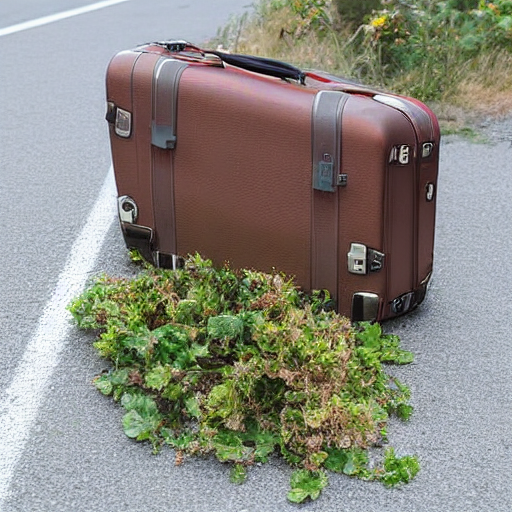}& \myim{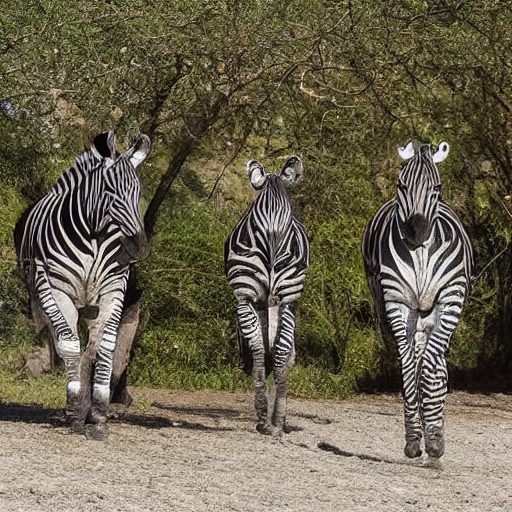}
\end{tabular}
\myvspace
\caption{Qualitative comparison of \ours to other methods based on LDM, conditioning on COCO  captions and up to three segments. 
}
\label{fig:qualitativeSupMat}
\end{figure*}

\section{Additional visualizations}
\label{sec:add_viz}
\mypar{Evolution of attention maps across timesteps} We show in \fig{attention_guidance_bear} and \fig{attention_guidance_fruit} average attention maps on the different objects present in the input segmentation during the first 12 denoising steps with and without our guidance scheme. We condition on the same Gaussian noise seed in both cases. 
We notice that attention maps quickly converges to the correct input conditioning mask when we apply \ours and that the attention masks are already close to ground truth masks only after 12 denoising iteration steps out of 50.

\mypar{Additional visualizations on COCO} In Figure \ref{fig:qualitativeSupMat}, we show additional qualitative samples generated with COCO masks comparing \ours to the different zero-shot methods.

\begin{figure*}
\centering
 \def\myim#1{\includegraphics[width=30mm,height=30mm]{#1}}
     \centering
   \setlength\tabcolsep{1pt}
   
   \renewcommand{\arraystretch}{0.2}
     \begin{tabular}{ccccc}
     \centering
       ``A \blues{car} and  a \reds{tree},  &`` A \reds{mirror}, \blues{sink} &  ``\blues{Plate with cookies} & ``A \blues{brown cow} in & ``A \yels{mouse} wearing \\
       at the beach.'' & and \yels{flowers} & and \reds{cup of coffee}, & a field, cloudy sky, & \blues{a hat} in the desert.'' \\
       & in a bathroom.''& fancy tablecloth '' & \reds{red full moon}'' \\
 \myim{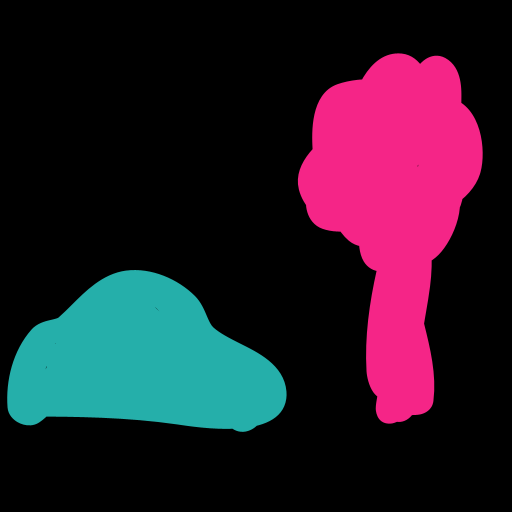} & 
\myim{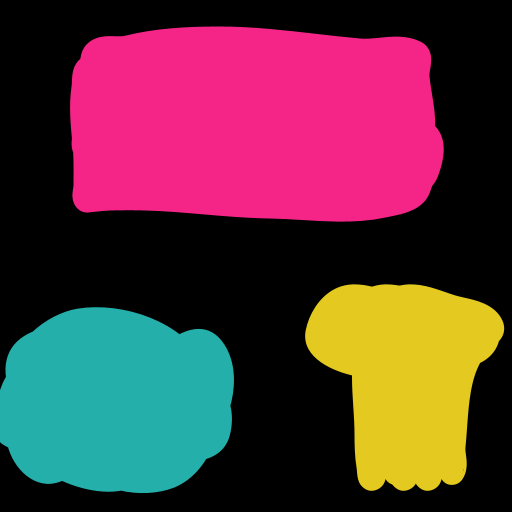} & 
\myim{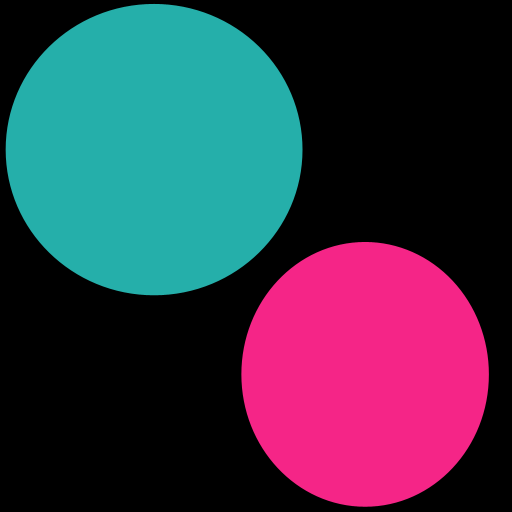} & \myim{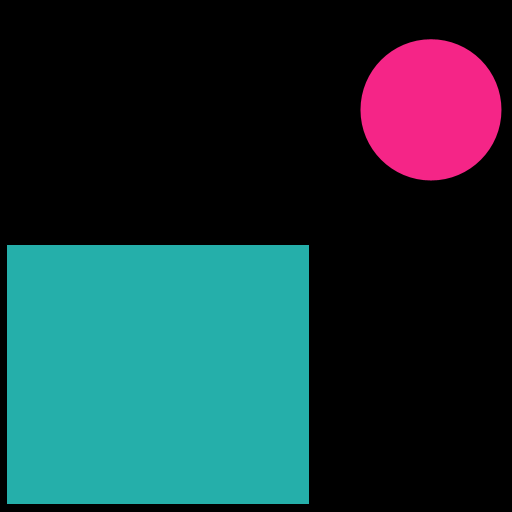} &
\myim{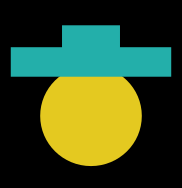}\\
 \myim{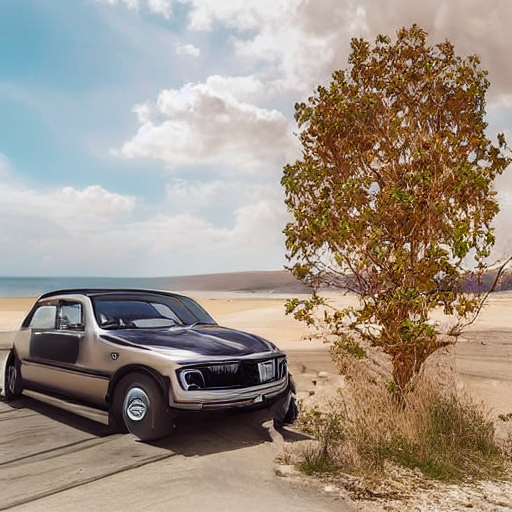} & 
\myim{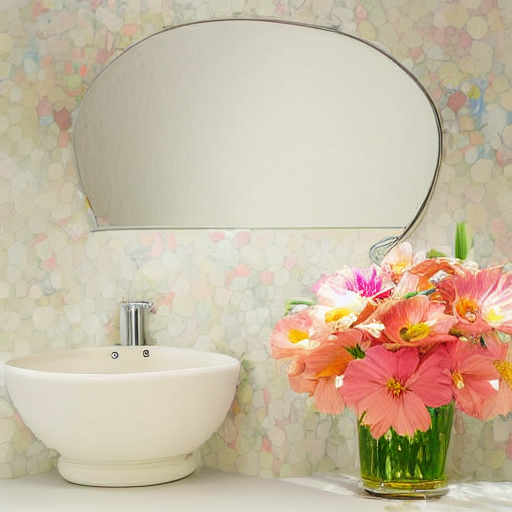} & 
\myim{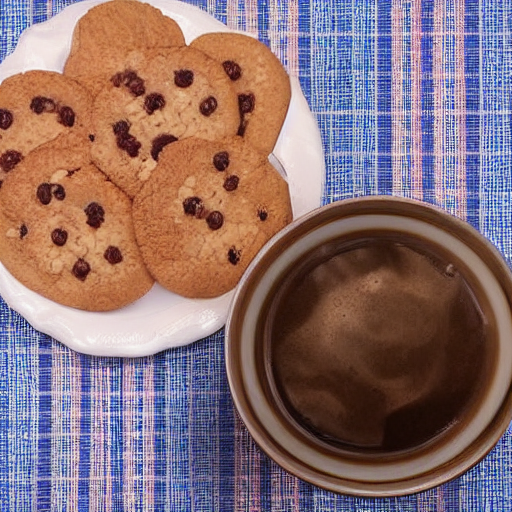} & 
\myim{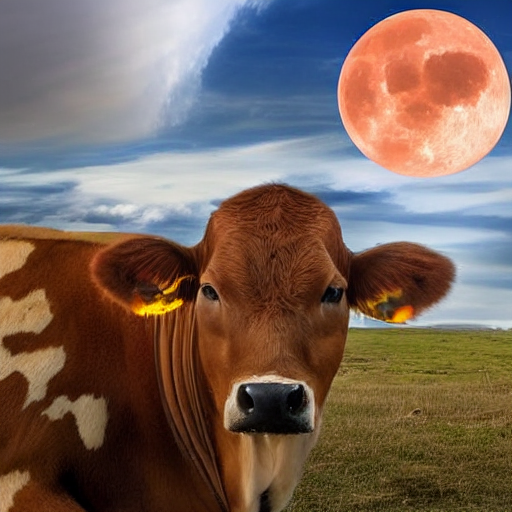} & 
\myim{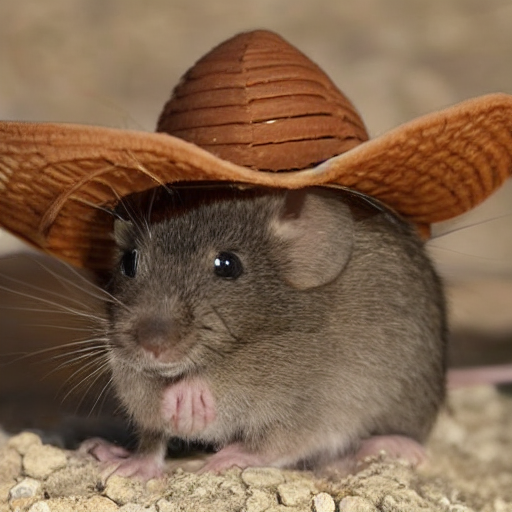}

\end{tabular}
\myvspace
\caption{\ours generations on coarse hand-drawn masks.}
\label{fig:quali_supmat}
\end{figure*}

\mypar{Visualizations on hand-drawn masks} In \fig{quali_supmat}, we show generations conditioned on coarse hand-drawn masks, a setting which is closer to real-world applications, similar to Fig.\ 2 in the main paper. 
In this case the generated objects do not exactly match the shape of conditioning masks: the flexibility of \ours helps to generate realistic images even in the case of unrealistic segmentation masks, see \eg the cow and mouse examples.

\end{document}